\documentclass[10pt, a4paper]{article}
\usepackage{lrec2022} 
\usepackage{multibib}
\newcites{languageresource}{Language Resources}
\usepackage{graphicx}
\usepackage{tabularx} 
\usepackage{soul} 
\usepackage{array,multirow,graphicx}

\usepackage{titlesec}
\titleformat{\section}{\normalfont\large\bfseries\center}{\thesection.}{1em}{}
\titleformat{\subsection}{\normalfont\SmallTitleFont\bfseries\raggedright}{\thesubsection.}{1em}{}
\titleformat{\subsubsection}{\normalfont\normalsize\bfseries\raggedright}{\thesubsubsection.}{1em}{}
\renewcommand\thesection{\arabic{section}}
\renewcommand\thesubsection{\thesection.\arabic{subsection}}
\renewcommand\thesubsubsection{\thesubsection.\arabic{subsubsection}}

\usepackage{epstopdf}
\usepackage[utf8]{inputenc}

\usepackage{hyperref}
\usepackage{xstring}

\usepackage{color}

\title{The ComMA Dataset V0.2: \\Annotating Aggression and Bias in Multilingual Social Media Discourse}

\name{\begin{tabular}{c}Ritesh Kumar\textsuperscript{2},Enakshi Nandi\textsuperscript{1},
Laishram Niranjana Devi\textsuperscript{1},
Shyam Ratan\textsuperscript{2}, \\
Siddharth Singh\textsuperscript{2}, 
Akash Bhagat\textsuperscript{3},  
Yogesh Dawer\textsuperscript{2}
\end{tabular}} 

\address{\textsuperscript{1}Panlingua Language Processing LLP, \textsuperscript{2}Dr Bhimrao Ambedkar University, \textsuperscript{3}Indian Institute of Technology-Kharagpur \\
         New Delhi, Agra, Kharagpur \\
         comma.kmi@gmail.com
         }

\abstract{
In this paper, we discuss the development of a multilingual dataset annotated with a hierarchical, fine-grained tagset marking different types of aggression and the ``context" in which they occur. The context, here, is defined by the conversational thread in which a specific comment occurs and also the ``type" of discursive role that the comment is performing with respect to the previous comment. The initial dataset, being discussed here (and made available as part of the ComMA@ICON shared task), consists of a total 15,000 annotated comments in four languages - Meitei, Bangla, Hindi, and Indian English - collected from various social media platforms such as YouTube, Facebook, Twitter and Telegram. As is usual on social media websites, a large number of these comments are multilingual, mostly code-mixed with English. The paper gives a detailed description of the tagset being used for annotation and also the process of developing a multi-label, fine-grained tagset that can be used for marking comments with aggression and bias of various kinds including gender bias, religious intolerance (called communal bias in the tagset), class/caste bias and ethnic/racial bias. We also define and discuss the tags that have been used for marking different the discursive role being performed through the comments, such as attack, defend, etc. We also present a statistical analysis of the dataset as well as results of our baseline experiments with developing an automatic aggression identification system using the dataset developed.
 \\ \newline  }

\begin{document}

\maketitleabstract

\section{Introduction} 

Aggression, bias, polarisation and hate are now commonplace phenomena on all kinds of social media platforms. And so are the research efforts to automatically identify these and process them in some meaningful way so as to reduce their harmful impact on social communication and society, in general. Two recent systematic surveys \cite{Vidgen2020} and \cite{Poletto2021ResourcesAB} have shown that over 60 datasets annotated with different aspects of hateful and abusive speech have been developed in various languages of the world in just around half a decade. While this rapid research is an encouraging sign, it has also led to some non-trivial issues. Some of these issues include (a) a bias towards resource-rich, largely European languages; (b) incompatible, sparse efforts and datasets (as pointed out by \cite{Vidgen2020} as well) and; (c) relatively less reliability of the annotation scheme and datasets (partially because of the lack of theoretical rigour and a somewhat “folk” approach to the problem). While the only way to promote research on more languages is to build datasets in more languages (we are making an effort towards that by including Meitei and Bangla in this dataset and we plan to include more languages as we move ahead), a possibility to alleviate the other two problems is to use what we call the 'discursive' methods of annotation for pragmatic phenomena like abusive and hateful language. Fundamentally the discursive methods argue for (a) building tagsets in a theoretically robust way, using ideas and concepts that have been thoroughly investigated in the field of pragmatics (so that we already have a good understanding of the phenomena that we are trying to annotate); (b) explicit declaration of the social and communicative background of the annotators as well as the overall ideological stand taken for the annotation of the dataset so as to make the rationale behind decision-making somewhat transparent; (c) documenting all the major decisions and points of discussion during the annotation process and making them publicly available along with the dataset such that they serve as guidelines for not just annotating new datasets but also for understanding the process of arriving at the annotations done in the dataset. This paper is a demonstration of this discursive method - both in terms of the annotation process as well as the documentation and reporting of this process.

In this paper, we discuss the development of dataset annotated with different kinds of aggressive language and bias in four languages viz. Meitei, Bangla, Hindi and English. We have used a modified version of the tagset discussed in \cite{kumar-etal-2018-aggression}. These modifications in the tagset are made considering the need to reduce the complexity of the earlier tagset and also in accordance with the needs of comprehensively annotating the available dataset, based on the observation of trends and patterns of discursive behaviour in each of these languages. For instance, the category of caste/class bias was included upon observing the nature of the comments in Bangla that were directed against the beggar-turned-singer Ranu Mondol, whose overnight success and subsequent change in attitude led to a deluge of comments directed not just at her gender but also at her lower class and caste identities. Similarly, the category for ethnic/racial bias was included upon observing the online animosity in the interactions between the Meitei and Kuki tribes in the Meitei data. We will discuss these decisions in more detail in the following sections of the paper.

This paper has been organized as follows. Section 2 gives a brief literature review on the frameworks and models that have been built to study the language of aggression and detect hate speech. Section 3 discusses the process of data collection and annotation. We also discuss the process and reasons motivating the modification of the earlier tagset and arrive at the current tagset. Section 4 discusses the annotation guidelines that have been developed and used for annotating the current dataset. Section 5 presents an analysis of the dataset and its limitations. We also briefly discuss the results of our baseline experiment for automatic aggression identification. Section 6 presents some concluding remarks on the tagset and guidelines thus developed.

\section {Previous Studies on Aggression}

The wide range of interrelated phenomena that has been used for annotating the datasets include broad dimensions such as abusive language \cite{Nobata2016}; \cite{waseem-etal-2017-understanding}, toxic language (\cite{Kolhatkar2020}; \cite{Kaggle2020}), aggressive language (\cite{Haddad2019}; \cite{kumar-etal-2018-aggression}; \cite{bhattacharya-etal-2020-developing}), offensive language (\cite{Chen2012}; \cite{mubarak-etal-2017-abusive}; \cite{Nascimento2019}; \cite{dePelle2016}; \cite{schafer-burtenshaw-2019-offence}; \cite{zampieri-etal-2019-predicting}, \cite{zampieri-etal-2019-semeval}; \cite{zampieri-etal-2020-semeval}), hate speech (several including \cite{Akhtar2019}; \cite{Albadi2018}; \cite{Alfina2017}; \cite{bohra-etal-2018-dataset}; \cite{Davidson2017}; \cite{malmasi-zampieri-2017-detecting}; \cite{schmidt-wiegand-2017-survey}), threatening language (\cite{Hammer2017}) or narrower, more specific dimensions such as sexism (\cite{Waseem2016}; \cite{waseem-hovy-2016-hateful}), misogyny, Islamophobia (\cite{chung-etal-2019-conan}; \cite{Vidgen2020Detectingweak}, homophobia \cite{Akhtar2019}, etc. Some of the datasets include a combination of these such as hate speech and offensive language (\cite{Martins2018}; \cite{mathur-etal-2018-detecting}) or sexism and aggressive language \cite{bhattacharya-etal-2020-developing}. In fact, \cite{jurgens-etal-2019-just} has rightly recommended for the abusive language researchers to broaden their scope so as to also study more subtle (such as condescension) as well as more serious forms of abuse (such as doxxing) and also posit the research within the broader framework of justice. 


However, despite the well-established practices of defining annotation schemes, providing detailed annotation guidelines and also reporting results of the inter-annotator agreement experiments for any task in NLP, a large number of researchers engaged in the abusive language research have chosen to ignore these. \cite{Poletto2021ResourcesAB} notes that “each resource is built referring to ad hoc definitions of the phenomena addressed, shaped so as to be suitable for a specific purpose, but what often lacks is a wider view on the topic and an eye towards interoperability of resources”. Furthermore, a large number of papers describing the datasets do not give any information about the guidelines that they have used for annotation nor the inter-annotator agreement score is mentioned (\cite{Davidson2017}; 
\cite{Ross2017}).


It could only be assumed from this that folk, common-sensical or simply convenient understanding of these phenomena are employed for preparing the datasets and training the model. It is generally accepted that these phenomena, viz. abusive, offensive, aggressive, toxic, hate(ful) and harmful language, are interrelated in some way but majority of the research rarely engages sufficiently well with their object of study to understand what exactly it is that they are trying to annotate and recognise. Some notable exceptions include \cite{waseem-etal-2017-understanding}; \cite{fortuna-etal-2019-hierarchically}; \cite{ousidhoum-etal-2019-multilingual}; \cite{Mandl2020} and \cite{kumar-etal-2020-evaluating}. \cite{waseem-etal-2017-understanding} posits a generic abusive language typology which may be utilised for annotating datasets according to broad categories of whether it is targeted or not and whether it is explicit or implicit - the typology is useful across annotating a wide range of abusive phenomena but may not very useful in understanding the interrelationships across those. \cite{ousidhoum-etal-2019-multilingual} puts different phenomena such as abusive, offensive, hateful, disrespectful and fearful as sub-categories of ‘hostility’ - however, there is no explicit attempt at establishing a hierarchical or scalar or any other kind of inter-relationship among these phenomena and are probably assumed to be independent of each other. On the other hand, \cite{Mandl2020} classifies hate speech, offensive and profanity as sub-categories of ‘hate and offensive’ - in this case also there is no explicit attempt at establishing their inter-relationship except stating that these are mutually exclusive, which may or may not be always true. 

One of the main reasons for using aggression as the meta-concept for annotating the dataset was our need to use a theoretically-motivated and well-researched phenomena for this study. We expected this to enable us to rigorously prepare the tagset, annotation guidelines and also analyse the dataset. Furthermore, using something a well-defined concept was also expected to be helpful in making the dataset more interpretable and interoperable. 

Aggressive language is listener-oriented, influenced by the surrounding and situational context of society, by community policy, social norms, social demographies, structural positions and relations, gender, age etc. It thus indicates the psychological, situational and structural perspective of the individual and the group of people \cite{weingartner-stahel-2019-online}. The comment or the post coming from the individual level becomes a matter of concern if it hurts the sentiments of a community or the ethnic group. The online platform with its diversity of user, their cultural differences and beliefs can be prove to be a catalyst for hate speech \cite{Al-Hassan2019}. Thus, online aggression on social media have the tendency to cause harm and becomes a source of creating  social problems.

In the present paper, we try to define and classify aggression in a theoretically-informed way, build a tagset out of it and use that for the automatic identification of aggression.



\section{Building and annotating an extensive dataset: Methodology}

Kumar, in \cite{kumar-etal-2018-benchmarking}, proposed a detailed classification and tagset for marking aggression and bias, which included the distinction between overt covert aggression as well as the target-based classification such as misogynistic, communal, geographical, sexual, etc.
While the tagset was quite detailed, it posed two problems - (a) there were too many tags to be comprehended and classified manually by annotators with an appropriate degree of precision; (b) it clubbed together non-mutually exclusive categories at the same level (for example, curse / abuse and non-threatening aggression were at the same level) - while this was handled somewhat by allowing for multi-label annotation, in principle, it was a non-rigorous scheme and posed several problems at the time of annotation.

In order to present the classification in a more principled way and also to practically ease the task of annotation, the scheme has been restructured and also extended to include those aspects which were left out in the earlier version of the tagset.
This modified tagset now includes a gradation of the intensity of aggression in a comment (physical threat, sexual threat, non-threatening aggression, curse / abuse) at every level, the discursive role of the given aggressive comment also include three new / modified roles (counterspeech, abet / instigate and gaslighting), and the addition of two new categories are added to tag biased speech targeted at individuals or social groups on the basis of their caste / class and ethnic /racial identities given in Table \ref{holistic_tagset}.

\begin{table}[!h]
\begin{tabular}{l|c|cc}
\cline{1-3}
\multirow{2}{*}{} & \multicolumn{3}{c}{\textbf{Aggression}}                   \\ \cline{2-3}
\textbf{Code} & \textbf{Aggression Level} & \multicolumn{1}{l}{\textbf{TAG}} \\ \cline{1-3}

1.1    & Overtly Aggressive   & OAG \\ \cline{1-3}

1.2 & Covertly Aggressive   & CAG  \\ \cline{1-3}

1.3  & Non Aggressive & NAG  \\ \cline{1-3}

1.4  & Unclear & UNC  \\ \cline{1-3}

\multirow{2}{*}{} & \multicolumn{3}{c}{\textbf{Aggression Intensity Level}} \\
\cline{2-3}

\textbf{Code} & \textbf{Attribute} & \multicolumn{1}{l}{\textbf{TAG}} \\ \cline{1-3}

1A.1.1    & Physical Threat   & PTH \\ \cline{1-3}

1A.1.2    & Sexual Threat   & STH \\ \cline{1-3}

1A.1.3  & Non-threatening Aggression & NtAG \\ \cline{1-3}

1A.1.4  & Curse/Abuse Aggression & CuAG  \\ \cline{1-3}

\multirow{2}{*}{} & \multicolumn{3}{c}{\textbf{Discursive Role}} \\ \cline{2-3}

\textbf{Code} & \textbf{Attribute} & \multicolumn{1}{l}{\textbf{TAG}} \\ \cline{1-3}

1B.1.1 & Attack & ATK \\ \cline{1-3}

1B.1.2 & Defend   & DFN \\ \cline{1-3}

1B.1.3 & Counterspeech & CNS \\ \cline{1-3}

1B.1.4 & Abet and Instigate & AIN \\ \cline{1-3}

1B.1.5 & Gaslighting & GSL \\ \cline{1-3}

\multirow{2}{*}{} & \multicolumn{3}{c}{\textbf{The Gender Bias}} \\ \cline{2-3}

\textbf{Code} & \textbf{Attribute} & \multicolumn{1}{l}{\textbf{TAG}} \\ \cline{1-3}

2.1 & Gendered Comments & GEN \\ \cline{1-3}

2.2 & Gendered Threats   & GENT \\ \cline{1-3}

2.3 & Non-Gendered Comments & NGEN \\ \cline{1-3}

2.4 & Unclear & UNC \\ \cline{1-3}

\multirow{2}{*}{} & \multicolumn{3}{c}{\textbf{The Religious Bias}} \\ \cline{2-3}

\textbf{Code} & \textbf{Attribute} & \multicolumn{1}{l}{\textbf{TAG}} \\ \cline{1-3}

3.1 & Communal Comments & COM \\ \cline{1-3}

3.2 & Communal Threats   & COMT \\ \cline{1-3}

3.3 & Non-Communal Comments & NCOM \\ \cline{1-3}

3.4 & Unclear & UNC \\ \cline{1-3}

\multirow{2}{*}{} & \multicolumn{3}{c}{\textbf{The Caste / Class Bias}} \\
\cline{2-3}

\textbf{Code} & \textbf{Attribute} & \multicolumn{1}{l}{\textbf{TAG}} \\ \cline{1-3}

4.1 & Casteist/Classist Comments & CAS \\ \cline{1-3}

4.2 & Casteist/Classist Threats & CAST \\ \cline{1-3}

4.3 & Non-Casteist/Classist Comments & NCAS \\ \cline{1-3}

4.4 & Unclear & UNC  \\ \cline{1-3}

\multirow{2}{*}{} & \multicolumn{3}{c}{\textbf{The Ethnicity / Racial Bias}} \\
\cline{2-3}

\textbf{Code} & \textbf{Attribute} & \multicolumn{1}{l}{\textbf{TAG}} \\ \cline{1-3}

5.1 & Ethnic/Racial Comments & ETH \\ \cline{1-3}

5.2 & Ethnic/Racial Threats & ETHT \\ \cline{1-3}

5.3 & Non-Ethnic/Racial Comments & NETH \\ \cline{1-3}

5.4 & Unclear & UNC  \\ \cline{1-3}

\end{tabular}
\caption{The ComMA Project Tagset}\label{holistic_tagset}
\end{table}

This dataset was built over multiple stages of the project, in concordance with the stages of development of the tagset. In the initial stages, the data was collected from online news websites and were tagged on the basis of two parameters: aggression and misogyny. In the second stage, this tagset was expanded to include communal bias, and the term ``misogyny" was altered to ``gender bias" so as to take into account bias directed at all genders (including transgenders) as well as people with different sexual orientations. The data was collected from the popular social media app YouTube, and included data in English, Hindi, Bangla, and Meitei.

This data was collected from YouTube, Twitter and Telegram after identifying specific videos and channels that attracted a great deal of hate speech and aggressive speech in the comments section, which were directed at women and minority religious groups, especially Muslims. The process of identifying these videos and channels involved typing keywords and hashtags related to controversial socio-political, religious, or cultural events in the recent and not-so-recent past, which were covered in one of the four languages mentioned above. The keywords used for searching the videos and the amount of raw data (number of comments) they yielded are included in Table \ref{sources_keyword}.


\begin{table*}[!h]
\begin{center}
\begin{tabular}{l|c|c}
\cline{1-3}
\multirow{2}{*}{} & \multicolumn{2}{c}{\textbf{Meitei}}                   \\ \cline{2-3}
\textbf{Sources: Videos/Channels} & \textbf{Search Keywords} & \multicolumn{1}{l}{\textbf{Number of Comments}} \\ \cline{1-3}

\multirow{2}{*}{} &  \#koubru   & 33  \\ \cline{2-3}

& \#lawaimacha & 1  \\ \cline{2-3}

Twitter & \#manipurdaCAB & 2  \\ \cline{2-3}

& \#manipurdaILP & 19 \\ \cline{2-3}

& \#meiteimuslim &  4  \\ \cline{1-3}

& \ triple talaq manung hutna  & 390 \\ \cline{2-3}

&  \ minister bishorjit nupi  & 348  \\ \cline{2-3}

&   \ ccpur nupi  &  138 \\ \cline{2-3}

&  \ paktabi diana & 107 \\ \cline{2-3}

&  \ non-manipuri  & 77 \\ \cline{2-3}

&  \ pangal nupi  &  320 \\ \cline{2-3}

&  \ utlou case  &  206 \\  \cline{2-3}

 &  \ manipur da CAB & 1118 \\ \cline{2-3}

&  \ Naga Accord & 304 \\ \cline{2-3}

 YouTube & \ Manipurdagi Meitei Furup mutpa & 59 \\ \cline{2-3}

& \ Potti Kappe & 59 \\ \cline{2-3}

& \ Pangal gi identity  & 668 \\ \cline{2-3}

& \ Muslim macha meitei na hatpa & 509 \\ \cline{2-3}

& \ Momoco gi thoudok & 66 \\ \cline{2-3}

& \ Koubru Conflict & 39 \\ \cline{2-3}

& \ Rani Sharma & 164 \\ \cline{2-3}

& \ Brinda Kanano & 87 \\ \cline{2-3}

& \ M.U Normalcy & 577 \\ \cline{2-3}

& \ ADC Bill & 115 \\ \cline{1-3}

\multirow{2}{*}{} & \multicolumn{2}{c}{\textbf{Bangla}} \\ \cline{2-3}
\textbf{Sources: Videos/Channels} & \textbf{Search Keywords} & \multicolumn{1}{l}{\textbf{Number of Comments}} \\ \cline{1-3}

  & \#SaveBangladeshiHindus  & 257 \\ \cline{2-3}

& \#TripurarJonnoTrinamool  & 452 \\ \cline{2-3}

& \#kutta & 42 \\ \cline{2-3}

& \#khanki & 19  \\ \cline{2-3}

Twitter & \#magi & 2  \\ \cline{2-3}

& \#sala & 66  \\ \cline{2-3}

& \#shuor & 17  \\ \cline{2-3}

& \#harami & 62  \\ \cline{2-3}

& \#hijra & 22  \\ \cline{1-3}

  & Police Files & 936 \\ \cline{2-3}
& Khela hobe & 1147 \\ \cline{2-3}
YouTube & Bengal Ram Navami & 687 \\ \cline{2-3}
& Nusrat Jahan baby & 964 \\ \cline{2-3}
& Ranu Mondol & 457 \\ \cline{2-3}
& Pori moni & 329 \\ \cline{1-3}

\multirow{2}{*}{} & \multicolumn{2}{c}{\textbf{Hindi \& English}}                   \\ \cline{2-3}
\textbf{Sources: Videos/Channels} & \textbf{Search Keywords} & \multicolumn{1}{l}{\textbf{Number of Comments}} \\ \cline{1-3}

 & BJP  & 848 \\ \cline{2-3}

Youtube & Ripped Jeans  & 903 \\ \cline{2-3}

& Nehru-Gandhi  & 947 \\ \cline{1-3}

& Hindutva & 2461   \\ \cline{2-3} 

Telegram & Love jihad & 249  \\ \cline{2-3}

& Feminism & 21  \\ \cline{1-3}

 & \#MuslimVirus & 1422  \\ \cline{2-3}

Twitter & \#BengalBurning & 651  \\ \cline{1-3}

\end{tabular}
\caption{Sources of Raw Data, Keywords and Number of Comments}
\label{sources_keyword}
\end{center}
\end{table*}




This dataset was then manually annotated by multiple annotators using a method named the `Discursive Methods of Annotation'. It has been demonstrated in several pragmatic and social science studies that the judgment of speakers on socio-pragmatic phenomena like aggression or bias (or even hate speech) is a function of:

\begin{itemize}
    \item \textbf{Contextual factors}, or more specifically the discursive experiences of the speaker, including what kind of discourses the speaker has been a part of, with whom, how many times, under what circumstances, and multiple other factors (see \cite{Agha2006} and his theory of enregisterment for some details on this). At the time of annotation the speakers continuously discuss their decisions with each other, whereby they modify each other's discursive experience(s) and possibly arrive at a space of mutual (dis)agreement. 
    \item \textbf{Co-textual factors} such as what else is included in the text as well as in its immediate context. 

\end{itemize}

At this point, the three co-textual factors included aggression, communal bias, and gender bias. While these are individually evaluated, they are also evaluated in the presence of each other during the process of annotation.

The socio-political position and ideological leanings of the annotators working on the dataset also play a significant role in determining how the data is analysed and tagged. While attempts are made after extensive discussions within the team of annotators to draw clear guidelines and definitions, supported by relevant examples from each language, for each tag and subtag so as to bring personal differences of opinion to a minimum, human differences between the socio-political, cultural, and ideological contexts of the annotators still manage to draw out some degree of difference in tags between the annotations of those working on the same dataset.

The details of annotators who worked on annotating the datasets is included in the Data Statement attached in Appendix I of the paper. 

During the course of annotating this dataset, we came across various other co-textual factors that triggered bias and aggression that had not yet been accounted for by our dataset. For instance, in the Bangla data, we observed caste and class based bias and aggression directed at the beggar-turned-singer Ranu Mondol, while in the Meitei data, we observed bias and aggression directed at various ethnic groups. This led to the inclusion of the categories of caste/class bias and ethnic/racial bias in the tagset, respectively.

Another major point of discussion that emerged from the annotation process was to do with the aggression tagset, which was divided into OAG (overtly aggressive), CAG (covertly aggressive), and NAG (non-aggressive speech). However, these sub-tags did not capture the range and intensity of aggression that was on display in the comments - from curse words to threats of a physical and sexual nature. This necessitated the introduction of a subtag under aggression titled Aggression Intensity Level (with tags for physical threat (PTH), sexual threat (STH), non-threatening aggression (NtAG), and curse/abuse based aggression (CuAG)), which would only be marked if the comment was tagged OAG or CAG.

Along with this, another optional subtag was added under Aggression, titled Discursive Effects, which is marked what the intended effect of a given comment is - attack (ATK), defend (DFN), counterspeech (CNS), abet/instigate (AIN), or gaslight (GSL). This particular subtag is used in a nuanced manner to help us identify the nature or discursive effect of comments that can be found on a thread, thus giving us an analytical tool to distinguish comments in threads from independent comments directed at the video or content creator. Each of the comments under a video is identified by a unique sequence of numbers, which help the annotators identify an independent comment from a thread in the spreadsheet they are working on. Using that knowledge, the discursive effects of a conversation in a thread are marked accordingly. It is to be noted that the annotators are discouraged from tagging comments as DFN, CNS, AIN, and GSL unless they feature in a thread, and are in response to a previous comment on the same thread. This particular subtag thus helps us distinguish the ways in which aggressive speech plays out when a commenter is engaging in a dialogue with another commenter in real time versus the kind of one-way conversations that characterize independent comments.    

This tagset is thus being developed during the course of annotating the raw data such that each is contributing to the development of the other. The new tagset is being used to annotate newer data, with the annotators conscious to highlight any shortcomings and flaws in the tagset that can be improved upon in the course of annotating the data, so that we can build a tagset that can be most optimal at identifying various forms of aggression and bias in social media interactions with the least margin for error.  

The complete tagset is given in Table \ref{holistic_tagset} and their definitions and examples are discussed in the following subsections. We are reproducing the definitions and examples of different levels of aggression and misogyny from \cite{bhattacharya-etal-2020-developing} since they have not undergone change since then.


\subsection{Aggression}

Aggression is classified on the basis of three broad levels, which have been discussed below with suitable examples. It is to be noted that we are annotating the way aggression is expressed in language but NOT the intensity of aggression expressed via the text at this level. There is a common tag for all categories, which is unclear (UNC) included with other tags in Table \ref{holistic_tagset}. 

\subsubsection{Overtly Aggressive (OAG)}
Any post/comment/text in which aggression is overtly expressed – either through the use of specific kinds of lexical items or lexical features which are considered aggressive and/or certain syntactic structures is to be annotated using this label. It includes direct attacks against the victim, and involves the use of commands, directives and specific kinds of lexical items such as curse words. Some examples of overt aggression are given below:

\begin{enumerate}
\item Abe Kanhaiya, suna inshallah (lal chaddi) gang Delhi MCD me 31 seat pe chunaav lada aur inhe kul 51 voton se azadi mili hai. Kamaal o gaya be.

 Oye Kanhaiya, I have heard that Insaallah (red chaddi) gang contested election on 31 seats in Delhi MCD and they got azadi (freedom) by a total of 51 votes. It is amazing.

\item BJP wale jyada dhindhora pitate h hindutva ka...aur hindu me hi equality nhi de pa rhe.isliye bjp ka virodh.baki states k compare me bjp ruled state me ye jyada hota h.isliye v.

The BJP people brag about Hindutva more than others and they are not able to ensure equality among Hindus. That is why this opposition against BJP. And also because this happens more in the BJP-ruled states.
\end{enumerate}

\subsubsection{Covertly Aggressive (CAG)}

Any post/comment/text in which aggression is not overtly expressed is classified as covert aggression. It
is an indirect attack against the victim and is often packaged as insincere, polite expressions (through the use of conventionalised polite structures), and includes sarcasm and satirical attacks. Some examples of covert aggression are given below:

\begin{enumerate}
\item Harish Om kya anti-national ko bail mil sakti hai?

Harish Om can an anti-national get bail?

\item PhD kab poori karoge? Kitne year tak scholarship loge, humlog k tax ka?

When will you complete your PhD? For how may years will you take scholarship out of our tax?
\item Beta teri ghar waapsi zaruri hai tu kuch zyaada hi shikshit ho gaya hai.

Son you need to return home, you have become a little too educated.

\item Ebungo ngse Meitei Ningol mchani kaonu... Pakistani mcha Muslim ntte ko… Don't used such wahei wata.

Don’t forget that you are the son of a Meitei woman… you are not a Pakistani Muslim… don’t use such words.

\end{enumerate}

\subsubsection{Non-aggressive (NAG)}
This label should be given to all those human speech samples which do not exhibit aggression in any form. Some examples of non-aggressive speech are given below:

\begin{enumerate}
\item Desh ka agla Jaiprakash..Agla Ambedkar yehi hai..

This is the next Jaiprakash and Ambedkar of the country.

\item Desh Ko tumhari jarurat hai…

The country needs you

\item Tum aage badho .. Hamari subhkamnaey Tumhare sath hai…

You move ahead..Our best wishes are with you.
\end{enumerate}

\subsubsection{Unclear (UNC)} 

In rare instances where it is not possible to decide how to tag a text displaying any kind of bias, it is tagged as unclear (UNC). Annotators are highly discouraged from using this tag and they are required to explicitly mark the text based on the other tags. However, in cases where this is indeed employed, all attempts will be made to resolve it through discussion and/or majority voting among multiple annotators. Under no circumstance will this tag make it to the final tagged document. It only serves as an intermediary tag for flagging and resolving really ambiguous and unclear instances in the annotation guidelines (discussed in \cite{bhattacharya-etal-2020-developing}).





\subsection{Aggression Intensity Level}
This category marks the intensity of aggression in the current post/comment. This feature is to be used only for aggressive posts/comments. As discussed earlier, all the sub-types for judging the intensity of aggression in the posts/comments discussed in \cite{kumar-etal-2018-aggression} are included in this version under aggression intensity level, which is illustrated in Table \ref{holistic_tagset}.


\subsubsection {Physical Threat (PTH)}
A post/comment is potentially physically aggressive (verbal aggression transforming into physical aggression) when it directly threatens to physically harm, hit  or even kill someone (an individual or the community). Some examples of physical threat in comments are:
\begin{enumerate}
\item abey saaley haarami teri aukaat kya hai tu hug dega aaja bihar tewra wIT KAR RAHA HU ....MAREGA  TU PAKKA .. JUB TAK DELHI ME HAI APNI KHUSIYA MANA MAI THOKUNGAA TERE KO

You moron, illegitimate son, what is your standing, you will pee in your pants. You come to Bihar, I am waiting for you. You will definitely die. You celebrate the time till you are in Delhi. I shall shoot you. 

\item Muh kala hai dogle ka dil bhi kala gaddar hai mujhe tum dikh jaye sala juta marunga dogala deshdrohi 
    
This hypocrite has lost his face, his heart is also bad, as soon as I shall see you moron, I will hit you with a shoe, you hypocritical anti-national.

\end{enumerate}

\subsubsection{Sexual Threat (STH)}
Any post/comment that contains words expressing sexual coercion and assault is marked as sexual threat.

\begin{enumerate}
\item Bhosri ke jab kuchh pata nahi hai to bolta kyu hai ja ke apni gaar marwa halala me.

You fucker, when you do not know anything then why do speak. Go and get yourself fucked in Halala.

\item Tok chudi

Let me fuck you

\item Momoco Fucker girl u looser fuck off
      
\end{enumerate}

\subsubsection{Non-threatening Aggression,(NtAG}
Any act of aggression targetted at personal attributes like one's intelligence, physical features, various identities, or anything else but not containing a threat is classified as non-threatening aggression. Examples include:

\begin{enumerate}
\item Bhut kam jankari h babu tmko.general caste ki aabadi 15 percent v kam h. general cat ka MATLAB hi h open category.

You know very little, boy. The population of general caste is less than 15 percent. The meaning of general category is open category. 

\item Tum sirf patar patar bolna jaante ho aur kuch nahi. Kuch acche karm bhi kar liya karo, gareeb dil se dua denge, sukuun milega tujhe. 

You only know how to blabber and nothing else. You do some good work also, people will bless you from their heart, you will get peace. 
\end{enumerate}

\subsubsection{Curse/Abuse Aggression (CuAG)}
Any comment containing an act of aggression that involves cursing or abusing the victim is tagged as curse/abuse aggression. Some examples of curse/abuse aggression are given below:
\footnote{All the examples given in this document are ‘real-life’ examples and they are reproduced exactly as they appear in the actual post/comment and form part of our corpus. The comments written in the Devanagari and Bangla scripts have been transcribed in the Roman script for the benefit of the readers.}

\begin{enumerate}
    \item Mujahid Irfan meri kaonsi pol khol Di iss ghade kanhaiya ne 
    
    Mujahid Irfan what has this ass Kanhaiya revealed about me? 
    
    \item Abe kutte.… Tere pichwade me jo suvar baithe hai voh bhi tuje sun nahi rahe hai saale to apni gandi jubaan bandh rakh .....koi bhi sun nahi  raha 
    
    Oye dog the pig in your ass is also not listening to you you moron. So keep your dirty mouth shut..nobody is listening to you.
    
    \item Manipur se kasuba kasubi gi makon natte khngbra nama napana nahei tamhandana warktra jatlo
    
    Manipur is not a place for prostitutes and gigolos. Did your parents gave you proper education? 

\end{enumerate}

\subsection{Discursive Role}
This category refers to the role of the current post/comment in the ongoing discourse. It could be one of five kinds: attack (ATK), defend (DFN), counterspeech (CNS), abet / instigate (AIN), and gaslighting (GSL). Attack, defend, and abet were included earlier in \cite{kumar-etal-2018-aggression}. Now, it has been extended to include counterspeech and gaslighting, and has combined abet and instigate to form the tag abet / instigate (AIN). Each of these tags have been given in Table \ref{holistic_tagset} and discussed in detail below:

\subsubsection{Attack (ATK)}
Any comment / post which attacks a previous comment / post is tagged as attack. It can only be tagged for an aggressive comment. Some examples of attack are given below:

\begin{enumerate}
    \item Muh kala hai dogle ka dil bhi kala gaddar hai mujhe tum dikh jaye sala juta marunga dogala deshdrohi 
    
    This hypocrite has lost his face, his heart is also bad, as soon as I shall see you moron, I will hit you with a shoe, you hypocritical anti-national
    
    \item tera samdhi mulayam singh yadav bolta tha rape koi badi galti nahi baccho se galti ho jaati hai. phaasi ki kya jarurat hain. pravachan aur updesh khatam kaha hotey hain. 100 chuhe khaakar billa chala haj ko.
    
    Your child’s father-in-law Mulayam Singh used to say that rape is not a big mistake and children make such mistakes. What is the use of hanging till death. The preachings and speech hardly ends. He is now pleading innocence after committing hundreds of crimes. 
    
    \item Eikhoisu ymna tukutchei cgumbase adubu haothu haiba gumba cde haina roise mikhrana touraga eikhoi dasu panba dudi nungai Jade makhoi na touriba lichat tuge matung Ena makhoi da haikh loire. Haothu gaina.....keino cbu
    
    We also don't like the nature but don't use the word Haothu (derogatory word used towards tribal community). You cannot blame whole of us for the wrongdoing by some people. Say it to them. What is this????

\end{enumerate}

\subsubsection{Defend (DFN)}
Any comment / post which defends or counter-attacks a previous comment / post is tagged as defend. The previous comment / post must be an attack and the current one should be in support of the victim. It could be both aggressive as well as non-aggressive, but it is imperative that it be in the same thread as the previous comment. Examples include:

\begin{enumerate}
    \item Kitna dukhi hai bhai tu, lagta hai teri pool kholdi kanhaiya ne. Agar tu jo ilzam uspe laga raha hai wo sach hai toh wo kiyon jail me nahi hai. (Covertly aggressive) 
    
    How sad you are bro. It seems that Kanhaiya has shown your true face. If the accusation that you are levelling on him is true then why is he not in prison. 

    \item Av tak chargesheet file nhi kar payi h Delhi police...Aur lab ne is bat Ko confirm kiya ki anti-national slogan me Kanhaiya ki aawaz nhi h.ye dusre logo ne kiya.kon kiya hoga ...Samjhte hi hoge. (Non-aggressive) 
    
    Delhi Police has not been able to file chargesheet till now. And the lab has confirmed that the anti-national slogan does not contain the voice of Kanhaiya. It has been done by other people. You must have an idea who has done it. 
\end{enumerate}

\subsubsection{Counterspeech (CNS)}
Counterspeech is any direct response to hateful or harmful speech which seeks to undermine it. Just as influential speakers can make violence seem acceptable and necessary, they can also favorably influence discourse through counterspeech. The most direct way it can succeed is to have a positive effect on the speaker, convincing him or her to stop speaking dangerously now and in the future. It can also succeed by having an impact on the audience – either by communicating norms that make Dangerous Speech socially unacceptable or by ‘inoculating’ the audience against the speech so they are less easily influenced by it. There are two types of counterspeech: organized counter-messaging campaigns and spontaneous, organic responses (from Counterspeech | Dangerous Speech Project). Here, we will tag both kinds of counterspeech as CNS. Counterspeech is always non-aggressive in tone and content. Examples of each case is given below:

\begin{enumerate}
    \item Main bolna chahta hun Sabhi se a ki jo galat Karen use Gali do pure bhartiyon ko Gali na do please nahin to Sabhi Ke Munh Mein Gali aati hai aur aata hai aapko bhi Dena chalu kar diya Jaega isliye Jo galat Karen uske sath sajae ho ja ham log mante Hain vah karo Jo Sabke hit mein ho aap unke Upar Ishara Karke pure bhartiyon ko Gali de rahe ho ya galat kar rahe ho please yah video ko Sudhar ke bolo Gali dekar mat bolo please. 
    
    I want to tell everybody that you must abuse those who do wrong, but please do not abuse all Indians, otherwise everyone will abuse and it will be directed at you too. That's why we ask that those who do wrong must be punished. Do what we feel is right, do what is good for everybody. You are targetting (him) and abusing all Indians, this is wrong. Please improve this video and don't abuse, please.    
    
    \item Please do not defame Nehru. His personal life was his private affair. Judge him by his contribution to the country.
    
    Please do not defame Nehru. His personal life was his private affair. Judge him by his contribution to the country.
    
\end{enumerate}

\subsubsection{Abet / Instigate (AIN)}
Any comment / post which supports or encourages a previous (negative) comment and  instigates an individual or group to perform an aggressive act is tagged as abet/instigate. The difference between abet and instigate lies in how the speech relates to the act of aggression. Instigation happens before the event and its purpose is to trigger or provoke an act of aggression. Abetting is speech that occurs during or after the act of aggression and its purpose is to praise, support, and / or encourage that act as well as other such acts in the future. In many cases it may be difficult to distinguish whether a comment is abetting or instigating an act of aggression, as praise or support (abetment) for an act of aggression can be interpreted as instigating such acts in the future. The purpose of both is to enable and validate aggressive speech and actions, hence the two have been included under one category in this tagset.
\begin{enumerate}
    \item Great sachchai likha aapne
    
    You have written a great truth. 

\end{enumerate}

\subsubsection{Gaslighting (GSL)}
Any comment/post that seeks to minimize the trauma or distort the memory of a trauma faced by another person (usually mentioned in the previous post/comment) is tagged as gaslighting. It is not the same as offering consolation when someone expresses their pain or traumatic/negative experience. Gaslighting is a form of manipulation; it seeks to present things in a better light than it actually is. The intention of gaslighting is to make the previous poster / commenter to feel like what they are feeling, seeing, hearing, and saying is wrong, a misunderstanding, an exaggeration, an overreaction, or attention seeking behavior. 

One way to identify gaslighting is: it is always targeted at comments that talk about one's personal experiences (including traumatic and painful memories). Gaslighting looks very much like fact-checking at times and can be difficult to separate from the latter. It should only be marked in cases where it is clearly gaslighting. While it is possible to gaslight comments that present verifiable facts, unless it is very clear that it is in fact gaslighting, it is best not to tag it as such. Examples include:

\begin{enumerate}
    \item Tar cheye bhalo garur mangsho khao tomra. Tomra napak jinish bokkhon korcho. Tomra opobitro. Tomaderke pobitro korar cheshta korteche. Huzoor tomader bhalo chacche.

    (This is in response to a previous comment that says that the Maulvi referenced in the video is asking people to stop drinking cow's urine) It's better that you guys have cow's meat (than cow's urine). You people are consuming impure things. You are impure people. (The Maulvi) is trying to make you pure. Huzoor wants the best for you.  

    \item Ei Huzoor Sharia protishthar kotha bolche, e Hindu der humki dicche na. Aar hya, eti jene rakhun kono Muslim Hinduder otyachar kore na, borong Hindu ra Muslim der otyachar kore. Aapni Kashmir er dike takiye dekhen Hindura Muslimder otyachar korche, aapni Delhir dike takiye dekhen Hindura Muslimder otyachar korche, Muslimder masjid dhongsho korche.
    
    This Huzoor is talking about Sharia law, he is not threatening Hindus. And yes, you should know that no Muslim tortures Hindus, rather Hindus torture Muslims. Look at Kashmir, look at Delhi, they are destroying the Muslims' masjids.
    
    \item Amar Qarni wakhal du da ne ngaorise mi na ngang li ba mayam si kari leige khanda ba makhei eikhoigi di alah hairadi loire haina sum tou leiradi mi loinaba wagada ba manli. 
    
    Amar Qarni the mind of yours is confused because you are not ready to hear anything what the other person is trying to say but you are in into 'Allah is everything' concept.If you still believe Allah can do everything than no good will happen.
\end{enumerate}

The intention of the first comment is to manipulate the readers into believing that what the Maulvi, Huzoor, is asking for is for the good of the people, when it is actually a call to declare a Muslim sect, the Kadiyanis, as un-Muslims due to their "un-Muslim" beliefs and practices. 

The second comment intends to draw a picture that paints Muslims as eternal victims and Hindus as eternal torturers without taking into account the whole picture and acknowledging that both communities have harmed the other at various points in history. It intends to alter the cultural memory of historical events, and thus qualifies as a classic example of gaslighting. 

The third comment is a reply to earlier comment of portraying Islam as the best religion of all and no other people of different religion will understand. The intention is to express that every Muslim has the mindset of portraying "Allah as the best of all" and they should not forget that they are in Meiteiland.

\subsection{Gender Bias}
Any post / comment expressing bias based on the gender identity of its target comes under gender bias. It includes bias based on gendered stereotypes, traditional gender roles of the speaker or addressee, and biased references to one’s sexual orientation (discussed in \cite{bhattacharya-etal-2020-developing}). This tagset also includes Gendered Threat (GENT) and Non-Gendered (NGEN) comments. It has been given in Table \ref{holistic_tagset} and discussed in detail below:


\subsubsection{Gendered Comments (GEN)}
Any text expressing bias based on the gender identity or stereotypical gender roles of the speaker or addressee is tagged as GEN. It also includes biased references to one’s sexuality and sexual orientation, homophobic or transphobic slurs, and comments of a sexual nature . Examples include: 

\begin{enumerate}
    \item tere ma se puch sale tera bap kon h 
    
    Go and ask your mother who your father is.
    
    \item Shk ka d nupi cgi huithu ga mnle
    
    The face of the woman looks like hui(dog)-thu(vagina) (dog-vagina). 
    
    \item Rate ktw magir
    
    What is the rate of this slut?
    \end{enumerate}

\subsubsection{Gendered Threats (GENT)}
Any comment that can be read as a gendered threat to the well-being of the victim (such as rape threats or threat of physical violence owing to their gender role / identity) can be tagged GENT. Some examples of comments tagged with gendered threats are given below:

\begin{enumerate}
    \item She must be sent to some ranga billa first. Those who think these are funny names must read about these guys first before calling yourself by such names. 
    
    \item Nupi do hatok khro mahik mapini. 
    
    Kill the woman because she is the mother of all viruses.
    
\end{enumerate}

\subsubsection{Not Gendered Comment (NGEN)}
Any post / comment which does not display any gender bias will be marked non-gendered. Examples include:
\begin{enumerate}
    \item  Magise chinpha faraga lakpado kaothokhibagine twrise 
    
           She came memorising the line and forgot the lines.
    
           \item Muhammad was a demoniac person like this maula.
    
\end{enumerate}
The first comment is a sarcastic comment but tagged as NGEN as it is not gendered bias. 

The last comment is communal in nature but does not display gender bias, hence it has been tagged NGEN.

\subsection{Communal/Religious Bias}
Any post/comment that targets the real/presumed religious affiliation/identity/beliefs of the victim is tagged as communal (COM). These comments can be aimed against attempts to establish religious harmony and cooperation, can
discriminate against religious practices, propagate false religious ideas, or (negatively) stereotype an individual/community based on their religious identity.

The labels for tagging communal/religious bias is displayed in Table \ref{holistic_tagset}. These annotation guidelines are discussed below in detail.


\subsubsection{Communal Comments (COM)}
Any post/comment that targets the real/presumed religious affiliation/identity/beliefs of the victim is tagged as communal (COM). These comments can be aimed against attempts to establish religious harmony and cooperation, can
discriminate against religious practices, propagate false religious ideas, or (negatively) stereotype an individual/community based on their religious identity. Some examples of communal comments are given below:

\begin{enumerate}
    \item Is nalayk ko koi ye bataye ki jaha bhi Muslim bahut jyada ho jate h waha hinduo ko kyo nahi jine dete Pakistan or bagladesh me hindu ka kya hal h kashmiri pandit bisthapit kyo h karana or godhara me hinduo par atyachar kisne kiya 
    
    Somebody tell this idiot that wherever there is a large number of Muslim, why are Hindus not allowed to survive there. What is condition of Hindus in Pakistan and Bangladesh. Why were Kashmiri Pandits forced to migrate Who has victimised the Hindus in Karana and Godhara.
    
    \item (Don't bring your burrowed Christainity in the land of Kangleipak). Hill don't belong to tribal. Govt. didn't handover any land to any particular people.(Stupid people already destroy the forest even plant illegal poppy. History cannot be destroy by bringing refugees). Truth will always reveal one day. God will know who is doing the wrong things and the lies. They are getting all the benifit from the govt. and use the govt rule. (Then they will say the govt cannot control the hills area as hills belong to them.) The mind of the hill people are filled with hate and act as a enemy of all other community.... We Meetei is also tribal of this land. Accept the history. Accept the truth. (If they really concern about the environment they shouldn't plant the illegal poppy.. Everyone knows their evil mindset...) 
    
    The intention of the comment indicates a target towards the tribal community following Christianity. It targets the religious affiliation of the hill people. The comment also counts as overtly aggressive.
    
    \item Hoi adunabu huranba, acid attack amdi drug yonbagi chtnabi asi nongma2 ghenktlakeecdi.. Noi mohammad to rapist ne kwno, paedophile.. Seitan hido adune.. . Kasubi mcha
    
    For that reason only, drug case, acid attack cases are increasing. The Mohammad that you people believe in is actually a rapist, a paedophile and he is the Satan. Son of a bitch.

\end{enumerate}

\subsubsection{Communal Threats (COMT)}
Any comment/post that carries a threat to the well-being of the victim (such as threat of rape, riot, or physical violence) owing to their religious identity is tagged as a communal threat. Some examples of communal threat are given below:

\begin{enumerate}
    \item Dekho mere dalit bhaiyyo aap muslim ko support karte ho lakin muslim hamesha vistarvadi niti ke hote  Unke hisaab se koi dusra dharm nahi jee sakta Unka bus chaleto wo sabka gala katde tab to wo apko bi nahi chodenge Bus wo apka fayda utha rahe he 
    
    Look my Dalit brothers, you support the Muslims but the Muslims believe in being exclusive. They believe no other religion must exist. If it suited them, they'd slit everyone's throats. Then they won't spare you either. They are just taking advantage of you.

    \item  Echal eche singse takpise Islam gi mrmda.. Kari fttaba ywge hibado uthokse.. Nttrad manipur se khra leiradi kangleistan ongrani.. Meetei di mcha pokpsu hndre moidi machasu 6 tei tade.. Chekchlhwdrad akhoi se loina yotkhrani..Religion, civilization loina mngkhrni..Islamaphobia santokse
    
    Lets educate our sisters about the Islam and its wrong teachings. Else Manipur will become Kangleistan. They give birth to more than six children and Meetei are controlling birth. If we still ignore and not take necessary steps we will be eaten by them. Our religion and civilisation will vanish. SO, let us spread Islamphobia.

\end{enumerate}

\subsubsection{Non-Communal Comments (NCOM)}
Any post/comment that is not communal in nature will be tagged as non-communal. Examples include:

\begin{enumerate}
\item Awesome she is!!!

\item Musolmanra kono dhormoke chhoto korena.

Muslims do not cast aspersions on any religion.

\end{enumerate}

\subsection{Caste/Class Bias}

Any comment/post that targets the caste or class identity of the victim and discriminates against them on the basis of it is marked as caste/class bias. While caste and class are two distinct categories of social segregation that function in diverse and specific ways, they have been subsumed under one category in this tagset for two reasons. 

The first reason has to do with the  data which led to the addition of this category to the tagset. The Bangla data on Ranu Mondol contained comments directed at her low caste and class position in society. However, it was often difficult to discern which aspect of her class and caste identities was actually being targetted, as the comments were often derogatory but vague. It made sense to argue that the cumulative inferior position she occupied in society had partly triggered the vitriol against her (the others had to do with her conduct in the public eye); had she been only of lower caste or lower class, the nature of the comments might not have been much different. 

The other reason for clubbing caste and class together had to do with ease of annotation. The previous reason necessitated that we keep caste and class under one category to avoid mistagging a comment that was vague in nature.  An allied reason to support this decision was the need to keep the tagset compact  and prevent it becoming too extensive and unwieldy. Hence, in this tagset, caste and class have been  merged under one broad category. However, that does not mean that every comment tagged by this tag demonstrates both class and caste bias/ discrimination. In some cases it can flag caste bias, in others class, and in yet others, both.

These labels (Casteist/Classist Comments (CAS), Casteist/Classist Threats (CAST), Non-Casteist/Classist Comments (NCAS), and Unclear (UNC) have been shown in Table \ref{holistic_tagset}. 


\subsubsection{Casteist/Classist Comments (CAS)}
Any comment/post that targets the caste or class identity of the victim and discriminates against them on the basis of it is marked as caste/class bias. Examples include:

\begin{enumerate}
    \item Central govt k cabinet me dekho top k 10 ya 20  ministers ko.1-2 ko chor yahi 15 percent wale h.
    
    Look in the cabinet of central government. Among the top 10 – 20 ministers, besides 1 – 2, they are only these people with 15\% reservation.
    
    \item Uni matite boshe kheto na, uni dining table e boshe kheto. Kukurer pete ghee porle ja hoy aar ki
    
    She never ate on the floor, she ate at a dining table. This is what happens when you give a dog a taste of ghee.
\end{enumerate}
\subsubsection{Casteist/Classist Threat (CAST)}
Any post/comment that poses a threat to the victim based on their caste/class identity is tagged as Casteist/Classist Threat. 
\begin{enumerate}
    \item ei Ranu magi nijeke ki mone kore? Eke aar er meyeke chuler muthi dhore Ranaghat station e bhikarir bati dhoriye boshate hobe. Tokhoni giye era thik hobe.
    
    What does this Ranu think of herself? She and her daughter need to be dragged by their hair and made to sit in Ranaghat station with beggars' bowls. Only then will they be put in their place. 
\end{enumerate}

\subsubsection{Non-Casteist/Classist Comments (NCAS)}
Any post/comment that is not casteist or classist in nature is tagged as Non-Casteist/Classist. Examples include:
\begin{enumerate}
    \item Hindu sobai amra ek ekhane Brahman, Namashudra abar ki? Amra sobai Sanatani etai amader porichoy hok. 
    
    All of us Hindus are one, what is this Brahman, Namashudra etc.? We are all Sanatani, let this be our identity. 
    
    \item Seta lekha nai but ata dekte Hobe borno babosta k Kara koreche
    
    That is not written but we must see who made the caste system.
\end{enumerate}

\subsection{Ethnic/Racial Bias}
Any post/comment directed at and discriminating against the ethnic or racial or tribal identity, culture, language, skin colour, physical features, place of origin, nationality, etc. of the victim can be tagged as ethnic/racial bias. The annotation labels (Ethnic/Racial Comments (ETH), Ethnic/Racial Threat (ETHT), Non-Ethnic/Racial Comments (NETH) and Unclear (UNC)) are given in Table \ref{holistic_tagset}. 

\subsubsection{Ethnic/Racial Comments (ETH)}
Any post/comment directed at and discriminating against the ethnic or racial or tribal identity, culture, language, skin colour, physical features, place of origin, nationality, etc. of the victim can be tagged as ethnic/racial bias. Examples include:
\begin{enumerate}
    \item Kanglup Khudingmak hourakpham loina lei. Aduna Kuki singi hourakfam kaidano amta haibirak o.
    
    Every community has its own origin. So,please tell us the origin of Kuki.
    
    \item Ye chinkis ki bohat jal rahi he hamare desh ki Badti Military Power dekha kar. Sale China valo khud ke desh me original mal banana sikh lo pehle fir hume samjao.
    
    These chinkis (Chinese) are so jealous of the increasing military power of our country.You moron Chinese first learn to manufacture the original stuffs then teach us.
    
    \item Another in different places Tripuris were killed by Bengali.What about that?
    
\end{enumerate}

\subsubsection{Ethnic/Racial Threat (ETHT)}
Any post/comment that expresses the intention to inflict harm on an individual or community based on their ethnic or racial or tribal identity is tagged as ethnic/racial threat. Examples include: 
\begin{enumerate}
    \item Kuki, Naga and Meitei, all came to the soil of the present day Manipur much before Inida got independence. There is no way one can be termed refugee. Meitei and Nagas should think twice when they term one community as refugee otherwise, u r inviting bloodshed.
    
    \item Kuki Koubru touganu sigumba mathok oaktaba thabakse gyan tana hingmin na c mi amagi limit hyba amasu leibane madu keidw nungda pairage sanage kha ganu pokkani sathina sabani.
    
    Kuki people don't talk about Koubru.Lets live in a peaceful manner because everybody has a limit and if you are thinking to play with our patience then you will get burned. 
    
\end{enumerate}

\subsubsection{Non-Ethnic/Racial Comments (NETH)}
Any text which does not contain any ethnic or racial bias will be marked NETH.
\begin{enumerate}
    \item Amra shobai manush ager kotha bhebe luv nei amra mile thakte hobe tripura ke shundor korar jonno

    We are all humans, no point thinking about the past. We have to live in harmony to make Tripura a beautiful place.  
    
    \item It’s true. We should come out of communalism and live in peace. Think about it this way, we all have bad pasts but that doesn’t mean we cannot have a good future. Being educated we should stop ethnic riots. Ultimately, it’s human lives that are lost.
\end{enumerate}



It must be noted here that these tags of gender, communal, caste/class, and ethnic/racial bias are only used in comments that demonstrate some kind of bias against that subcategory. Mere mention of these social groups in a post/comment will not entail that they be tagged as such if there is no bias at play in the given post/comment. This point is amply at display in the examples furnished in the above sections. 

\section{Further Annotation Conventions}
Annotation based on this tagset is at the document level - it could be a comment or any one segment of the discourse, similar as (discussed in \cite{kumar-etal-2018-aggression} and \cite{bhattacharya-etal-2020-developing}). 
Further, this tagset is extended with the intensity of aggression and discursive role of any comment/post. Tagging of intensity is based on the use of aggressive vocab commentor used in comment/post. While, discursive roles are mainly marked where conversation is in thread as well as in normal comments/posts where it is looking like dialogue conversation.

\section{Inter-annotator agreement}
The first phase of the annotations was done by 9 annotators - 5 for Meitei, 3 for Hindi and 2 for Bangla. The Meitei, Bangla and Hindi has 140, 275 and 230 comments respectively. Comments are taken from YouTube and Twitter. The annotation is done using the aggression, gender bias and religious bias tagset. The Krippendorff's Alpha is used to measure inter-annotator agreement for the and Meitei,  Bangla and Hindi, given in Table \ref{iaa}.
We did not find any gendered comment in Meitei and communal comment in Bangla, thus the Table \ref{iaa} has not any IAA value for them. Overall from the Table \ref{iaa} we can say that the IAA is very less in first phase.

\begin{table}[!h]
    \begin{center}
    \small\addtolength{\tabcolsep}{-2pt}
    \begin{tabular}{c|c|c|c}
    \hline
    \textbf{Language} & Meitei & Bangla & Hindi \\
    \hline
    \textbf{Aggression} & 0.28 & 0.41 & 0.42 \\
    \hline
    \textbf{Gender} & - & 0.26 & 0.28 \\
    \hline
    \textbf{Religious} & 0.38 & - & 0.41 \\
    \hline
    \textbf{Number of Annotators} & 5 & 2 & 3 \\
    \hline
    \textbf{Number of Comments} & 140 & 275 & 230 \\
    \hline 
    \end{tabular}
    \caption{Inter-annotator agreement Phase 1}
    \label{iaa}
    \end{center}
\end{table}


Annotators were instructed to follow the annotation guidelines but keeping the option open for annotator's own interpretation while doing the annotation. The discussions based on the inter-annotator agreement experiment we have tried to reduce the disagreement among the annotators and reaching at an agreement in most of the instances. During the discussion sessions we felt need to add new categories and subcategories and tagset and updated the annotation guidelines.

\begin{table}[!h]
    \begin{center}
    \small\addtolength{\tabcolsep}{-2pt}
    \begin{tabular}{c|c|c|c}
    \hline
    \textbf{Language} & Meitei & Bangla & Hindi \\
    \hline
    \textbf{Aggression} & 0.51 & 0.66 & 0.93 \\
    \hline
    \textbf{Aggression Intensity} & 0.49 & 0.69 & 0.77 \\
    \hline
    \textbf{Discursive Role} & 0.66 & 0.74 & 0.75\\
    \hline
    \textbf{Gender} & 0.56 & 0.81 & 0.80 \\
    \hline
    \textbf{Religious} & 0.27 & 0.80 & 0.82 \\
    \hline
    \textbf{Caste / Class} & - & 0.49 & - \\
    \hline
    \textbf{Ethnicity / Racial} & 0.75 & - & - \\
    \hline
    \textbf{Number of Annotators} & 3 & 3 & 2 \\
    \hline
    \textbf{Number of Comments} & 205 & 208 & 209 \\
    \hline 
    \end{tabular}
    \caption{Inter-annotator agreement Phase 2}
    \label{iaaphase2}
    \end{center}
\end{table}

The results of Phase 2 of the inter-annotator agreement shown in Table \ref{iaaphase2}. It was conducted on 205 comments for Meitei, 208  and 209 comments for Bangla and Hindi respectively. There were 3 annotators in Meitei and Bangla  and 2 in Hindi. There is significant improvement in Phase 2 of IAA when compare to Phase 1. In all categories of Bangla (except Cast) and Hindi we get IAA of more than 65\%.
We did not find any casteist comments in Meitei and Hindi and ethnic comment in Bangla and Hindi in the dataset.

\section{Dataset}
\subsection{Training Set}
\begin{figure}[!h]
\centering
\includegraphics[width=1.0\columnwidth]{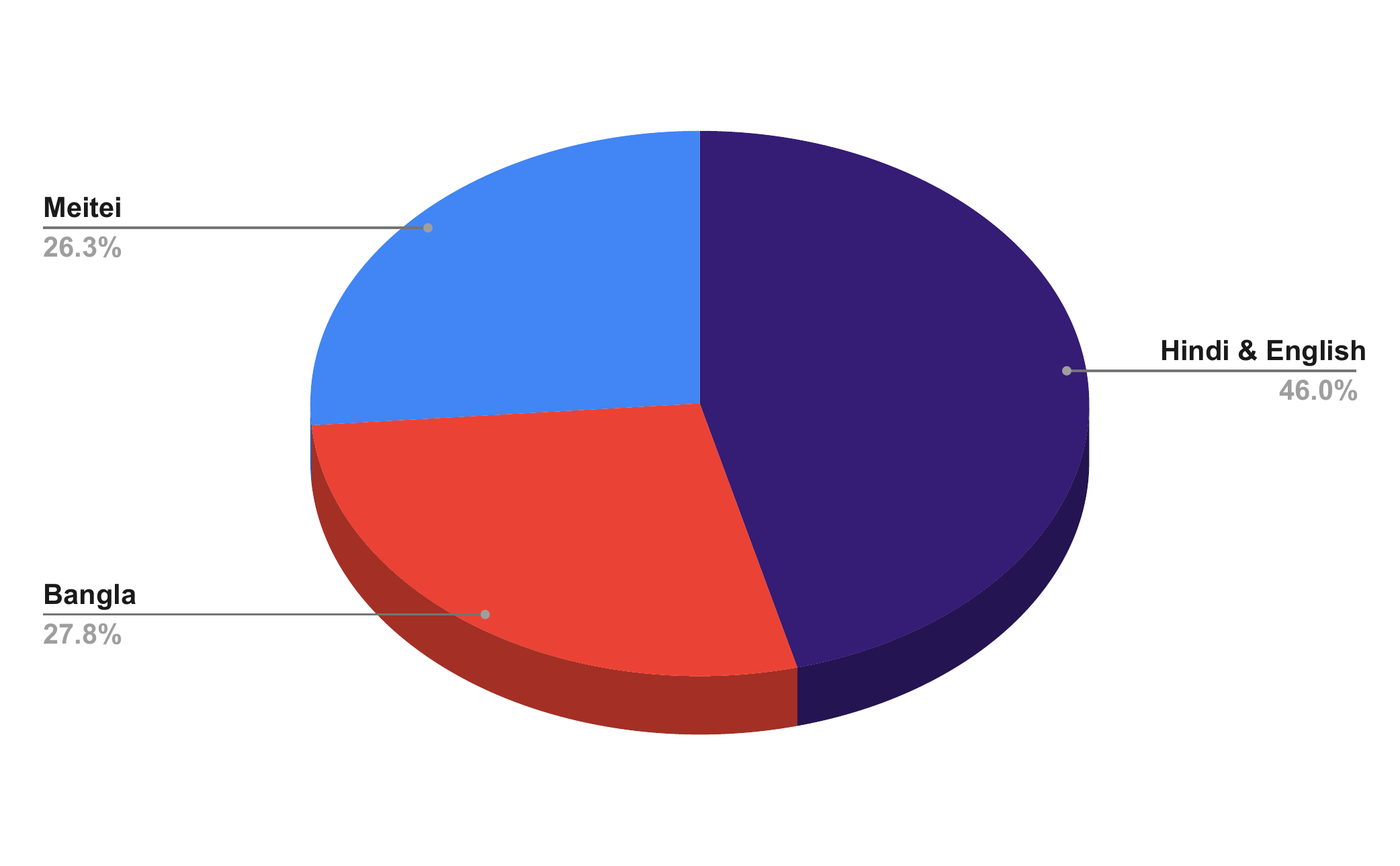}
\caption{Languages in the Dataset}
\label{lang}
\end{figure}

The training dataset contains a total of 12200 comments in four languages - Meitei, Bangla, Hindi and English.
Language-wise distribution is given in Figure \ref{lang}, which is precisely 26.27\% Meitei, 27.76\% Bangla,  45.97\% Hindi \& English.

\begin{table}[!h]
\begin{center}
\begin{tabular}{l|c|ccc}
\cline{1-5}
\multirow{2}{*}{} & \multicolumn{4}{c}{\textbf{Aggression}}                   \\ \cline{2-5}
& \textbf{TOTAL} & \multicolumn{1}{l}{\textbf{OAG}} & \textbf{CAG} & \textbf{NAG} \\ \cline{1-5}

\textbf{Meitei}    & \textbf{3,209}   & 456 & 1,495  & 1,258  \\ \cline{1-5}

\textbf{Bangla} & \textbf{3,391}   & 1,782 & 494  & 1,115 \\ \cline{1-5}

\textbf{Hindi \& English}  & \textbf{5,615}          & 3,052 & 969 & 1,594  \\ \cline{1-5}

\textbf{Multilingual}  & \textbf{12,211} & 5,289 & 2,956 & 3,966  \\ \cline{1-5}

\multirow{2}{*}{} & \multicolumn{4}{c}{\textbf{Gendered}} \\
\cline{2-4}

& \textbf{TOTAL} & \multicolumn{1}{l}{\textbf{GEN}} & \textbf{NGEN} \\ \cline{1-4}

\textbf{Meitei}    & \textbf{3,209}   & 203                            & 3,006 \\ \cline{1-4}

\textbf{Bangla}    & \textbf{3,391}   & 1,271                            & 2,120  \\ \cline{1-4}

\textbf{Hindi \& English}  & \textbf{5,615} & 1,175 & 4,440 \\ \cline{1-4}

\textbf{Multilingual}  & \textbf{12,211} & 2,647 &  9,564 \\ \cline{1-4}

\multirow{2}{*}{} & \multicolumn{4}{c}{\textbf{Communal}} \\ \cline{2-4}

& \textbf{TOTAL} & \multicolumn{1}{l}{\textbf{COM}} & \textbf{NCOM} \\ \cline{1-4}

\textbf{Meitei}    & \textbf{3,209}   & 242 & 2,967 \\ \cline{1-4}

\textbf{Bangla}    & \textbf{3,391}   & 416 & 2,975  \\ \cline{1-4}

\textbf{Hindi \& English}  & \textbf{5,615}         & 1,213 & 4,402 \\ \cline{1-4}

\textbf{Multilingual}  & \textbf{12,211} & 1,869 & 10,342 \\ \cline{1-4}

\end{tabular}
\caption{The ICON Training Dataset}\label{traindataset}
\end{center}
\end{table}

\begin{figure}[!h]
\centering
\includegraphics[width=\columnwidth]{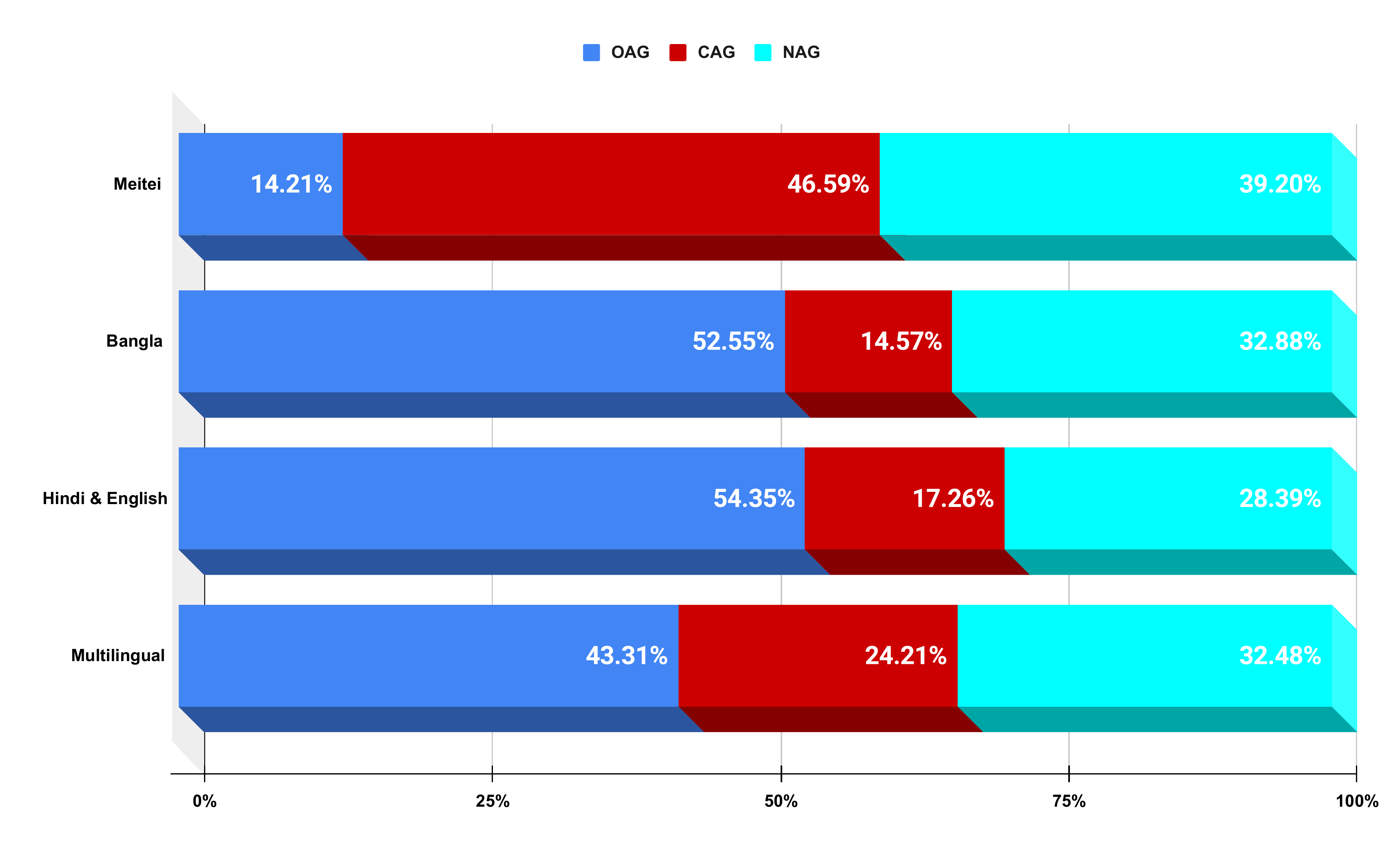}
\caption{Aggression in the Dataset}
\label{aggression}
\end{figure}

In Figure \ref{aggression} the distribution of overt, covert and non-aggression  in Meitei 14.21\%, 46.59\% and 39.20\%; in Bangla 52.55\%, 14.57\% and 32.88\%; and in Hindi \& English 54.35\%, 17.26\% and 28.39\%; respectively. The overall distribution of three major categories of aggression are 43.31\% overt, 24.21\% covert and 32.48\% non-aggression. In Meitei 46.59\% comments are covertly aggressive while in Bangla, Hindi and English more tan 50\% comments are overtly aggressive.

\begin{figure}[!h]
\centering
\includegraphics[width=\columnwidth]{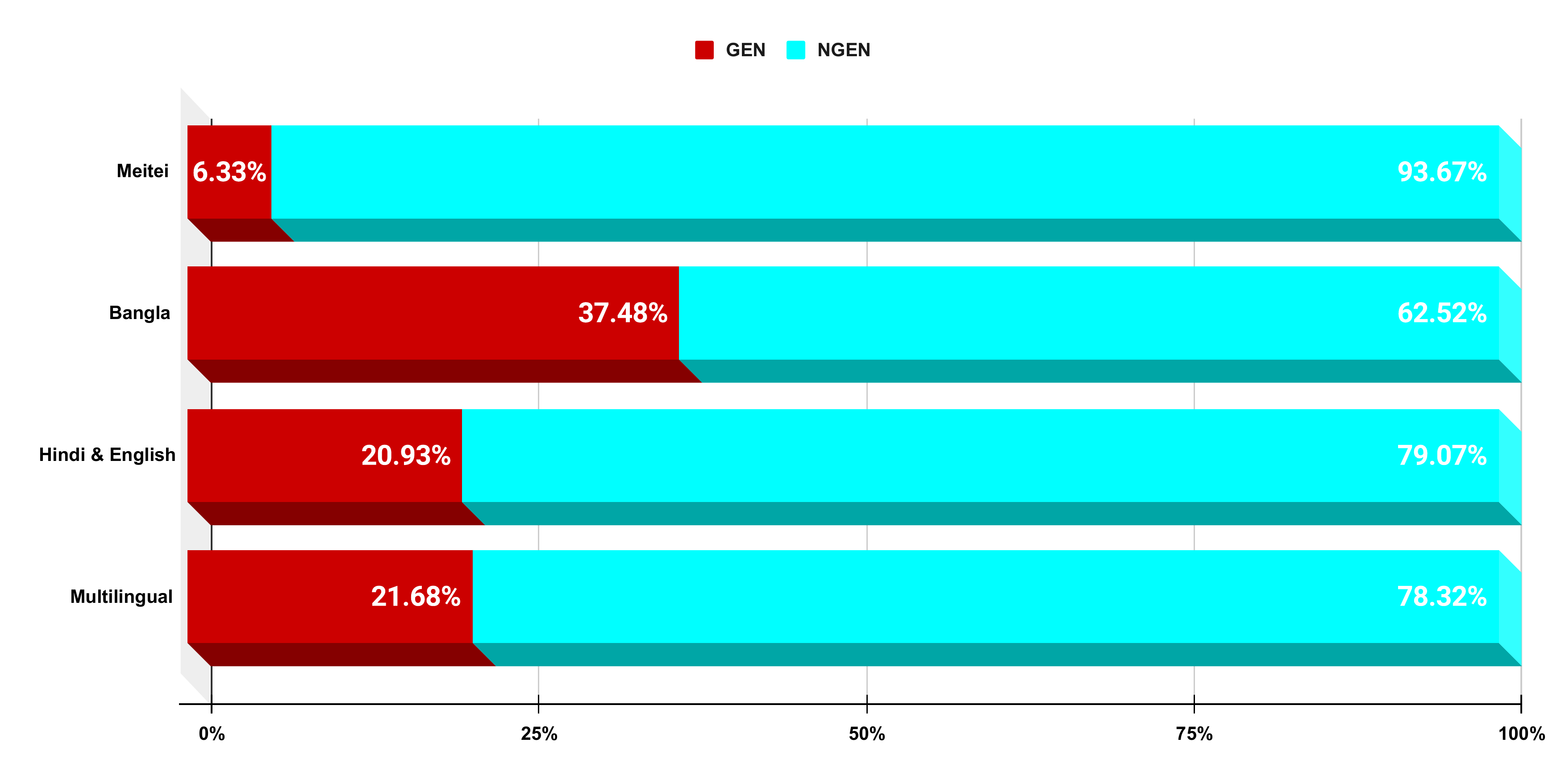}
\caption{Misogyny in the Dataset}
\label{gendered}
\end{figure}

\begin{figure}[!h]
\centering
\includegraphics[width=\columnwidth]{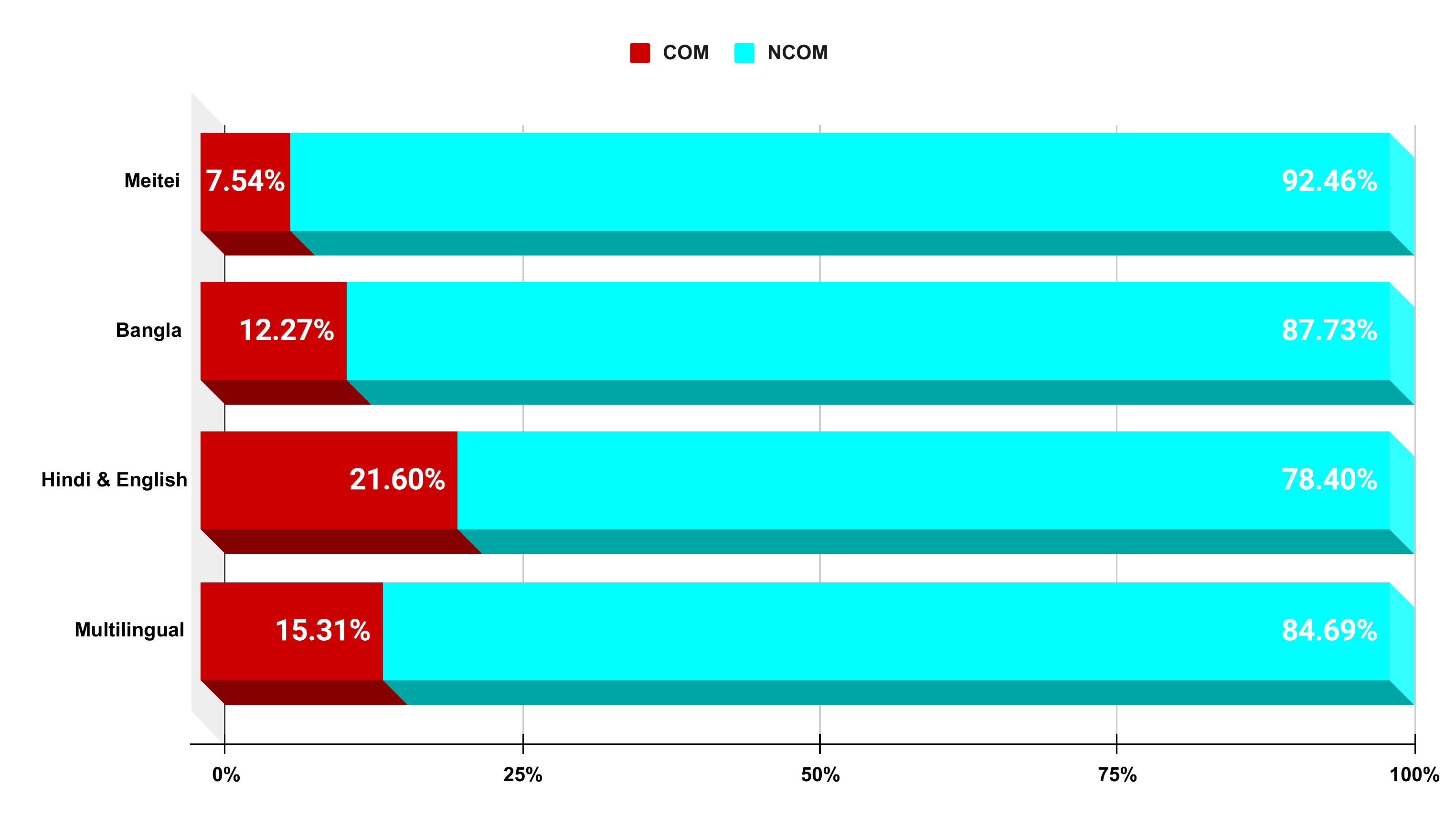}
\caption{Communal in the Dataset}
\label{communal}
\end{figure}

Overall 21.68\% portion of the comments are gendered and language-wise percentage in Meitei and Hindi is 6.33\% and 20.93\% respectively and 37.48\% for Bangla, shown in Figure \ref{gendered}.
Total communal comments are more than 15\% and language-wise communal comments distribution in Meitei, Bangla, and Hindi are 7.54\%, 12.27\% and 21.60\% respectively, given in Figure \ref{communal}. On broader scale Bangla has most number of gendered comments, whereas Hindi and English has most of communal comments.

\begin{figure}[!h]
\centering
\includegraphics[width=\columnwidth]{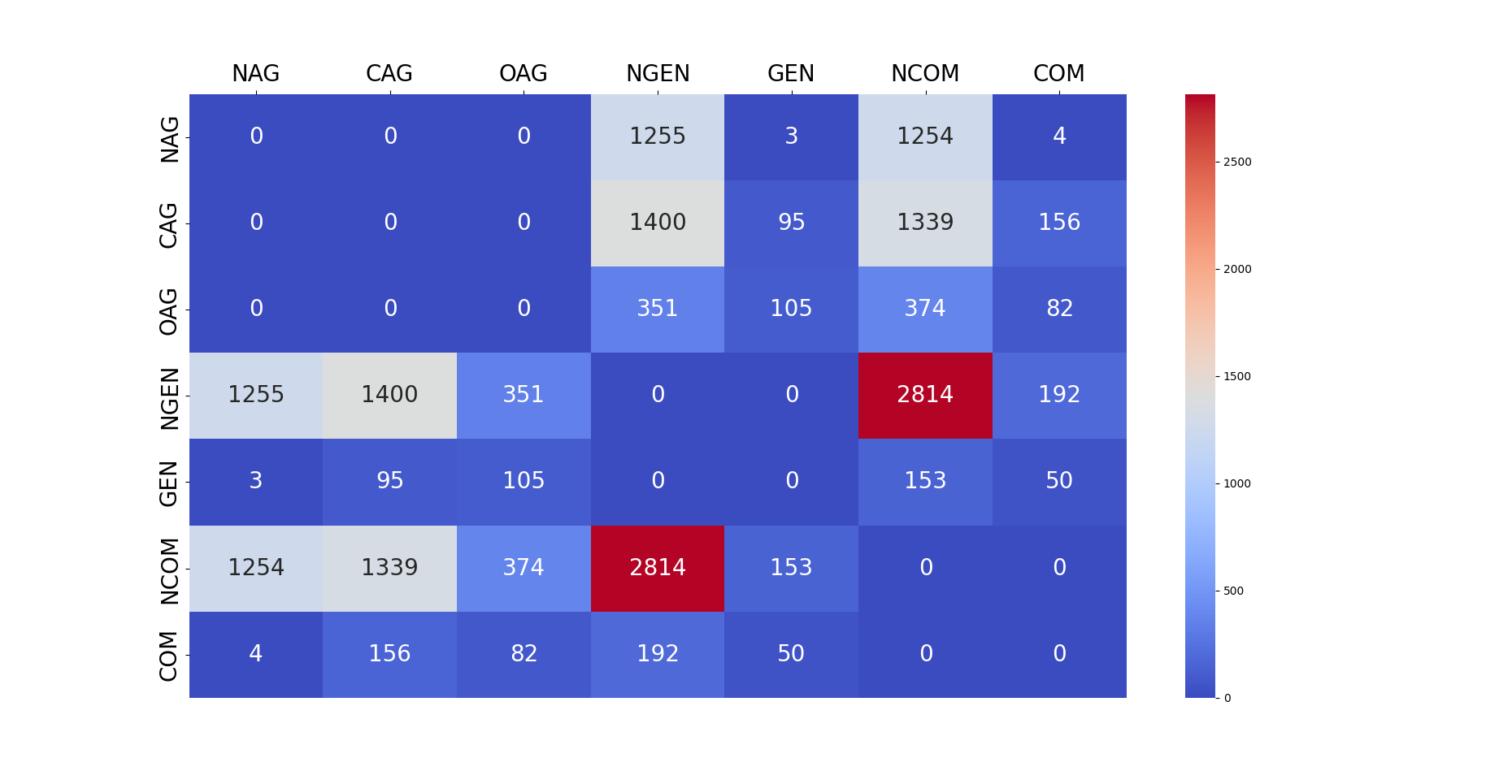}
\caption{Co-occurrence heatmap for Meitei Dataset}
\label{categorytuplemni}
\end{figure}

\begin{figure}[!h]
\centering
\includegraphics[width=\columnwidth]{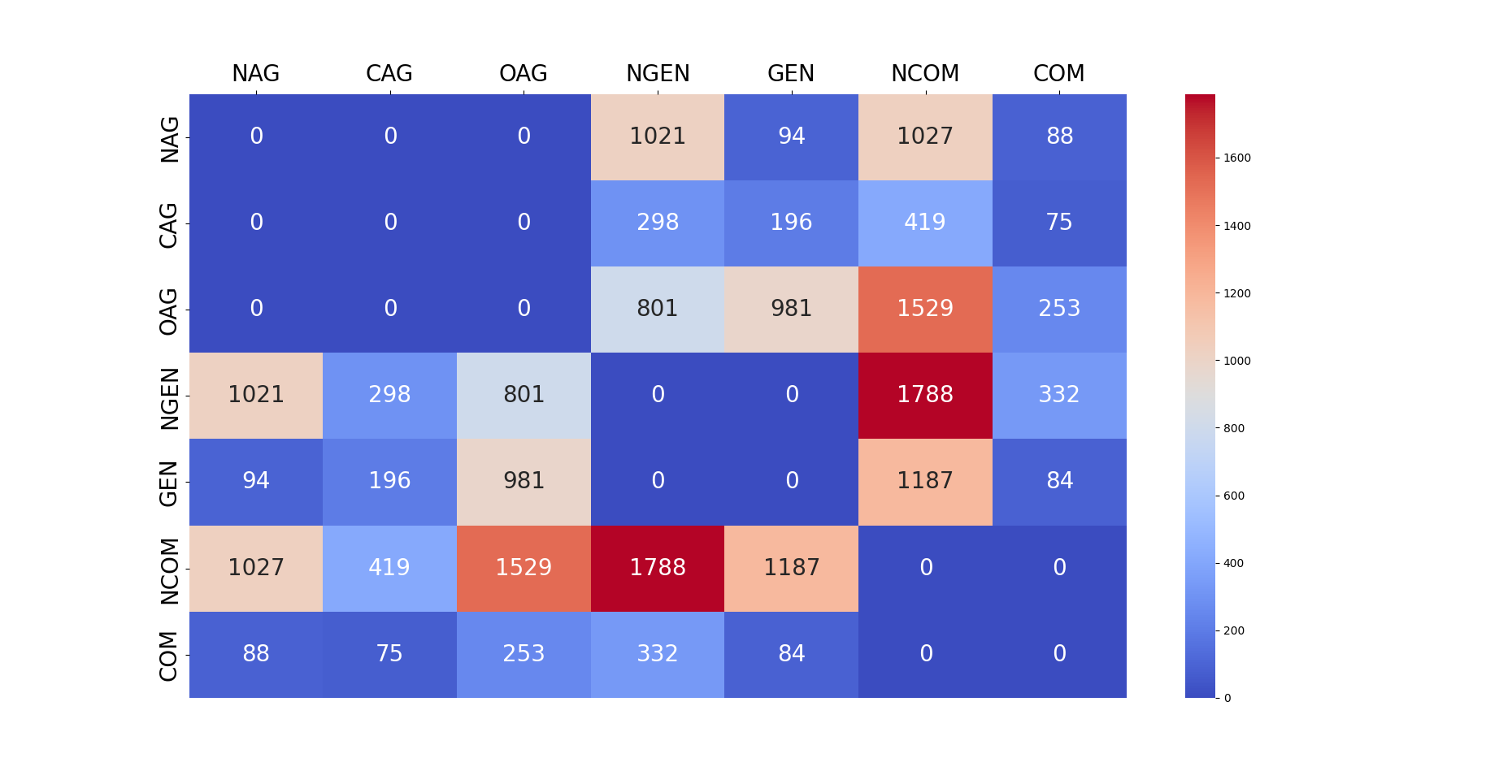}
\caption{Co-occurrence heatmap  for Bangla Dataset}
\label{categorytupleben}
\end{figure}

\begin{figure}[!h]
\centering
\includegraphics[width=\columnwidth]{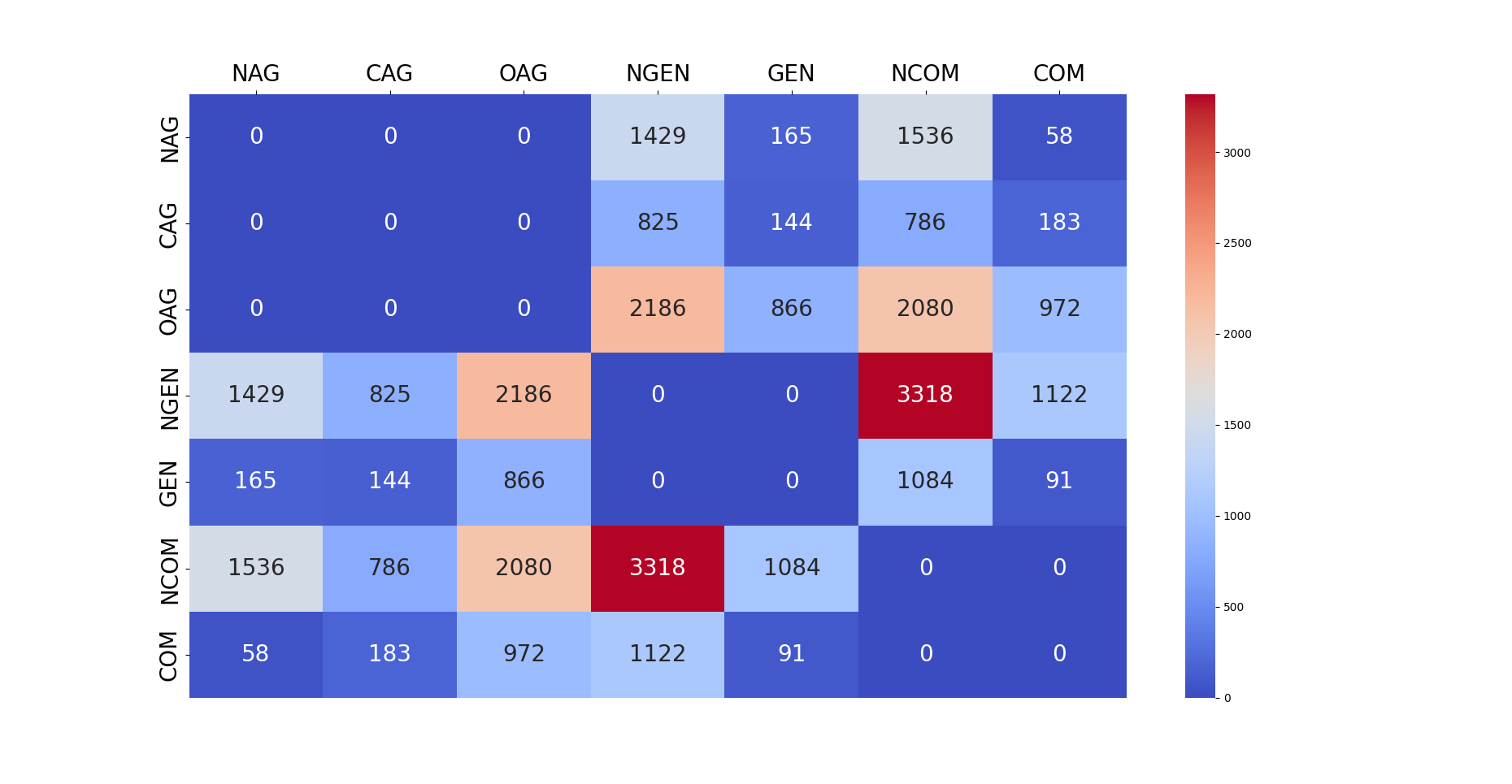}
\caption{Co-occurrence heatmap for Hindi Dataset}
\label{categorytuplehin}
\end{figure}

Co-occurrence graphs of Aggression, Misogyny and  Communal categories given in Figure \ref{categorytuplemni}, Figure \ref{categorytupleben}, Figure \ref{categorytuplehin} show that most of the communal comments are overtly or covertly aggressive in all languages. 
Similar pattern could be seen for gendered comments.

\begin{figure}[!h]
\centering
\includegraphics[width=\columnwidth]{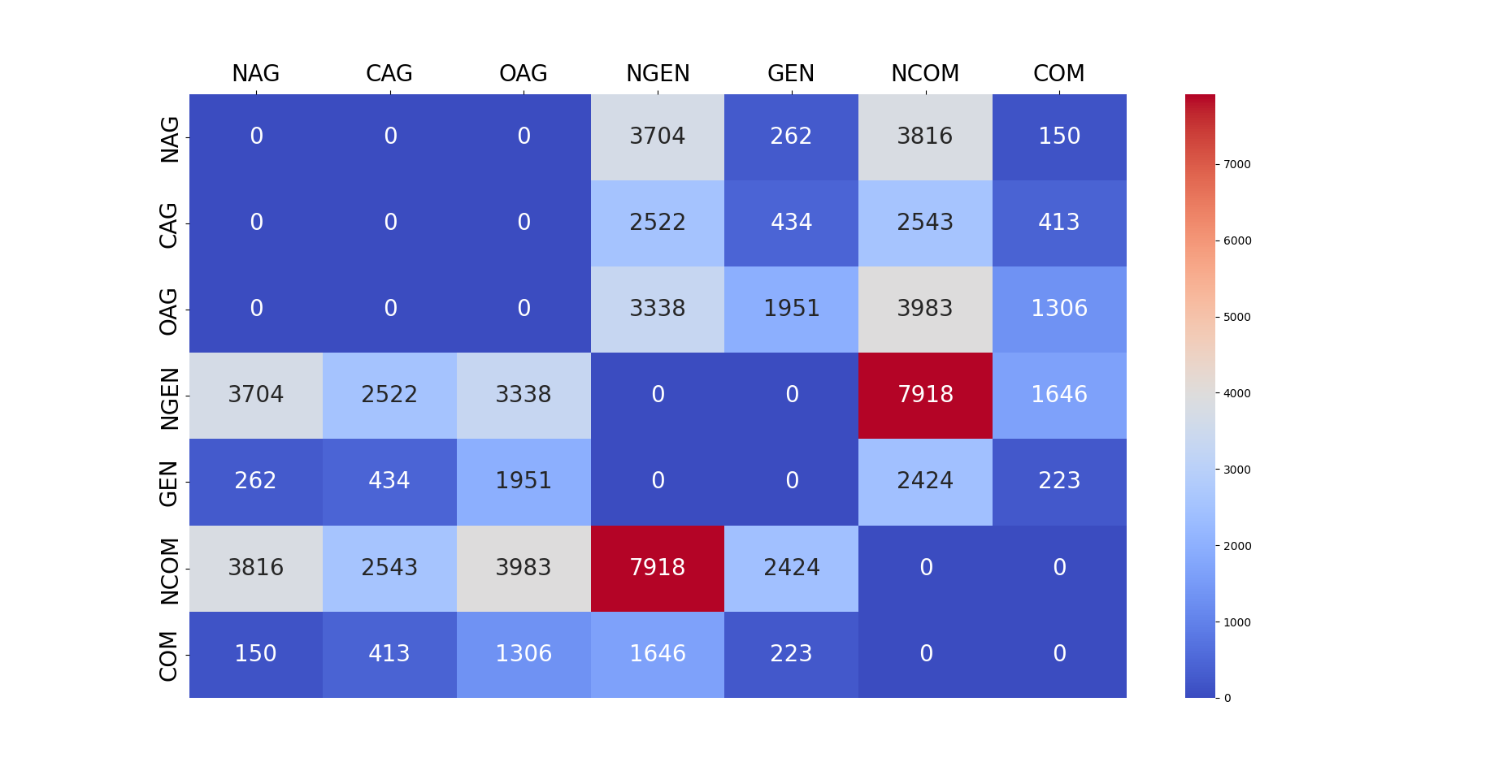}
\caption{Co-occurrence heatmap In The Dataset}
\label{categorytuplealllang}
\end{figure}

In the wider sense the distributions in co-occurrence graph in Figure \ref{categorytuplealllang}, we can say that if a comment is gendered or communal, possibility is that the comment has some kind of aggression.


\begin{table}[!h]
    \begin{center}
    \small\addtolength{\tabcolsep}{-2pt}
    \begin{tabular}{ m{2cm}| m{2cm} |  m{2cm} }
    \hline
    \textbf{Word} & \textbf{Frequency} & \textbf{Frequency Agg./Gen./Com.}  \\
    \hline
    \hline
    muslimvirus & 989 & 809 \\
    muslimsspreadingcorona & 146 & 122 \\
    india & 341 & 167 \\
    muslim & 338 & 178 \\
    meitei & 326 & 77 \\
    desh & 220 & 84 \\
    desh (in Devanagari) & 209 & 89 \\
    people & 193 & 92 \\
    hindu & 169 & 75 \\
    nupi & 166 & 83 \\
    covid & 163 & 128 \\
    modi & 191 & - \\
    muslims & 184 & 153 \\
    nehru & 173 & - \\
    pangal & 266 & - \\
    vote & 193 & - \\
    ram & 165 & - \\
    congress & 163 & - \\
    manipur & 257 & - \\
    bjp & 795 & - \\
    corona & - & 105 \\
    corona (in Devanagari) & - & 105 \\        
    maa & - & 100 \\
    women & - & 87 \\
    movie & - & 85 \\
    coronajihad & - & 82 \\
    randi & - & 81 \\
    teri & - & 81 \\
    \end{tabular}
    \caption{Union of Top 20 Most Frequent Words In The Dataset And In Aggressive/Communal/Gendered Or Combined Comments In The Dataset}
    \label{topfreqwords}
    \end{center}
\end{table}

When we do a union of the 20 most frequent words taken from comments which are marked aggressive/gendered/communal or combination of these and 20 most frequent words in the dataset we get the words shown in Table \ref{topfreqwords}. 
The words  like `muslim', `hindu'\footnote{Major religious group in India}, `covid', `congress'\footnote{Indian National Congress is a second major political party in India}, 'corona', `bjp'\footnote{BJP is a Hindu Nationalist Organisation currently heading the central government of India}, `vote' are mostly use in the context of religion and politics were present in communal comments.

The words like `maa'(mother), `randi'(prostitute), `nupi'(girl), `women' were used in the comments which are misogynistic and gendered, the word `teri'(your) frequently occurred with these words.

The comments in which the above mentioned words occurred, also turn out to be aggressive.
Most gendered/communal comments are aggressive in the dataset.

When we look at the count of the common words (the word which has some count in both the columns \textbf{`Frequency'} and \textbf{`Frequency Agg./Gen./Com.'}) and if we take the word `muslims', it has occured 184 times in the dataset out of which it has occured 153, about 83\% times in aggressive/gendered/communal comments.
If we take the word `hindu', it has occured 169 times in the dataset out of which it has occured 75, about 44\% times in aggressive/gendered/communal comments.
Similarly, the word `nupi'(girl) has occured 50\% of the times in aggressive/gendered/communal comments.
The words like `modi', `nehru', `ram', `congresss', `bjp' which do not have any count in the \textbf{`Frequency Agg./Gen./Com.'} column , they mostly occur in non-aggresive/non-gendered/non-communal comments.
The words like `maa'(mother), `women', `coronajihad', `randi'(prostitute), which do not have any count in the \textbf{`Frequency'} column , they mostly occur in aggresive/gendered/communal comments.


\begin{figure}[!h]
\centering
\includegraphics[width=1.0\columnwidth]{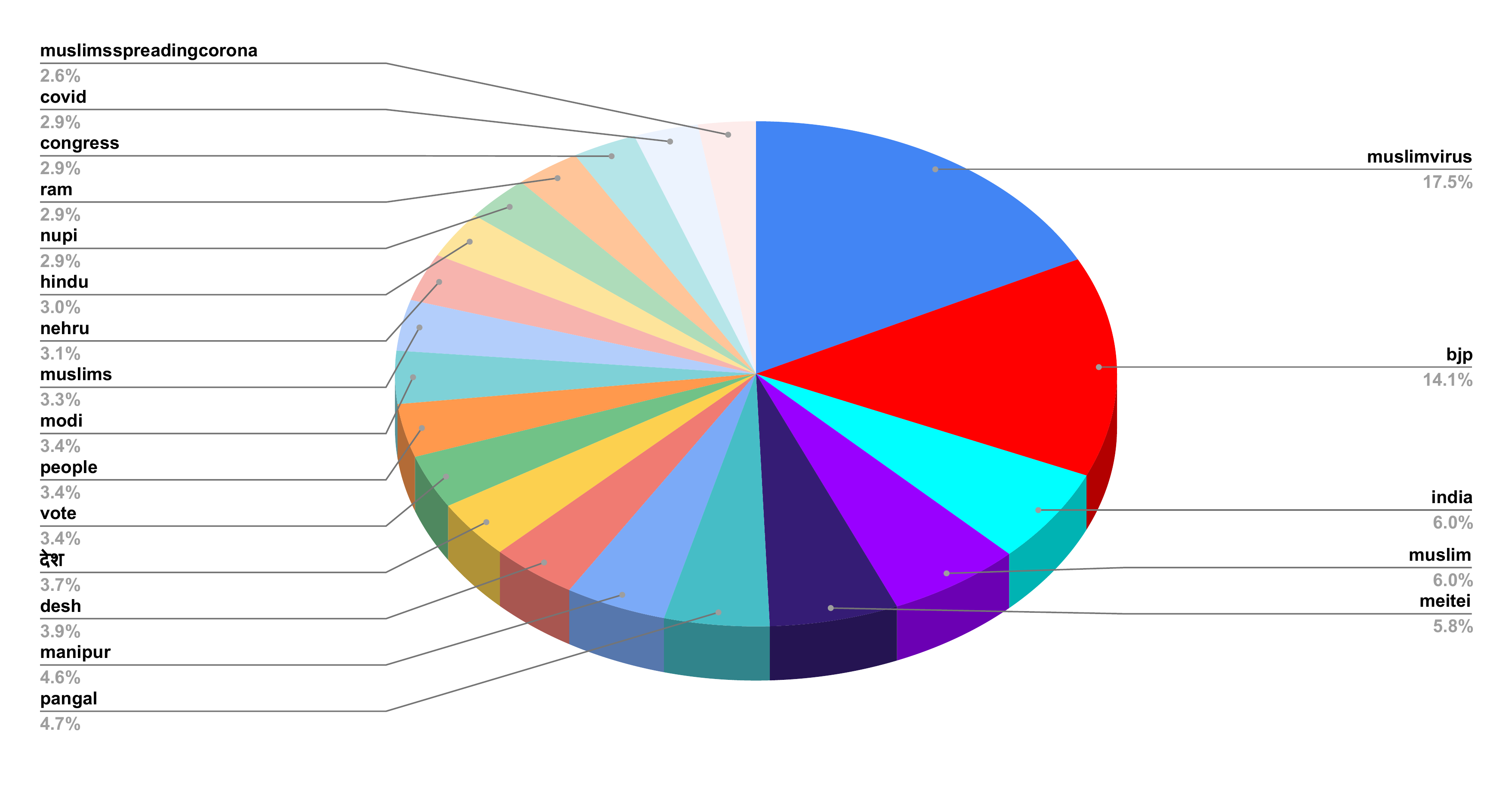}
\caption{Top 20 Most Frequent Words In The Dataset}
\label{top_frequency}
\end{figure}

\begin{figure}[!h]
\centering
\includegraphics[width=1.0\columnwidth]{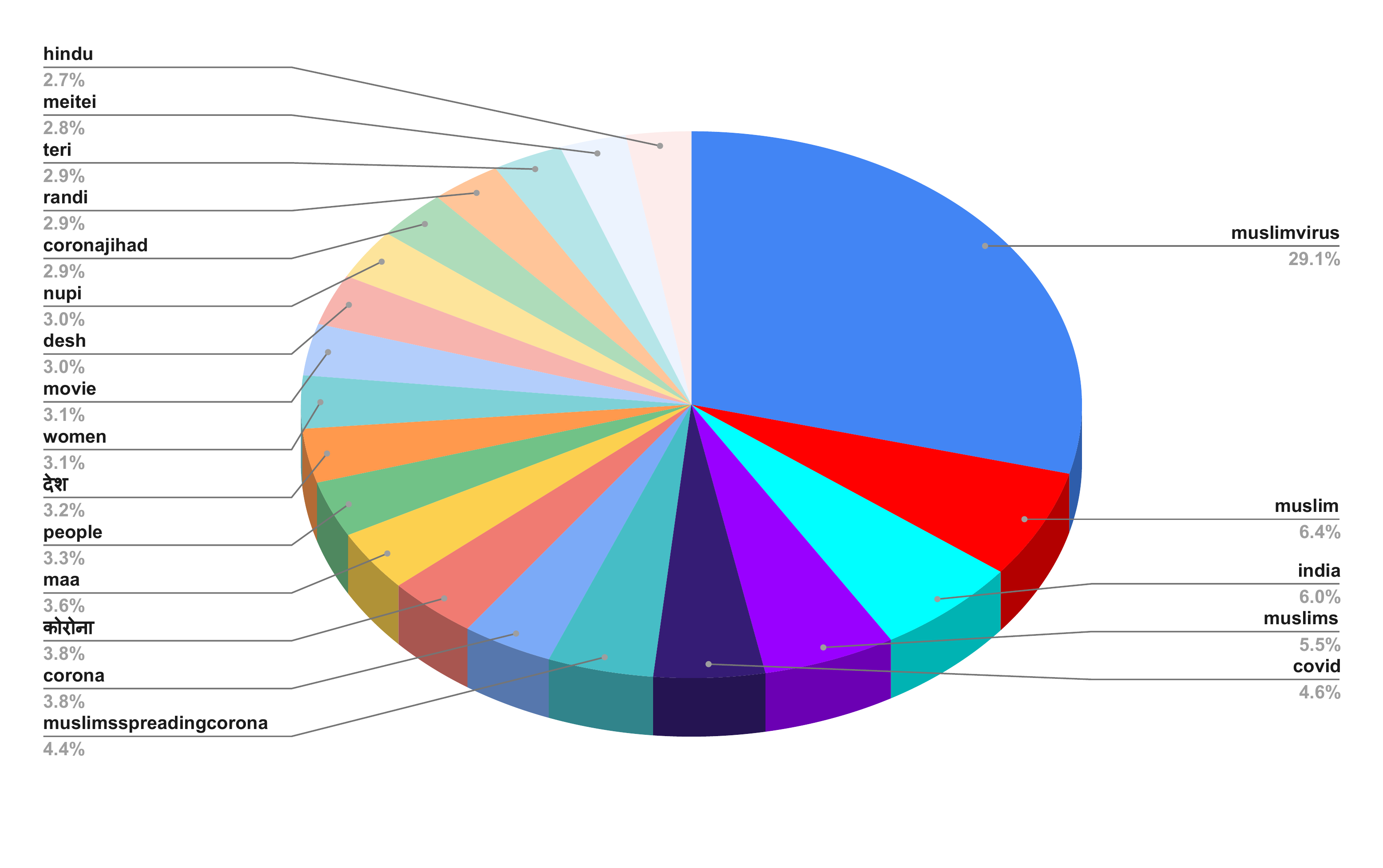}
\caption{Top 20 Frequent Words In The Dataset For Comments Which Are Either Aggressive/Communal/Gendered Or Combination Of These}
\label{top_combinedfrequency}
\end{figure}

\subsection{Test Dataset}

\begin{figure}[!h]
\centering
\includegraphics[width=1.0\columnwidth]{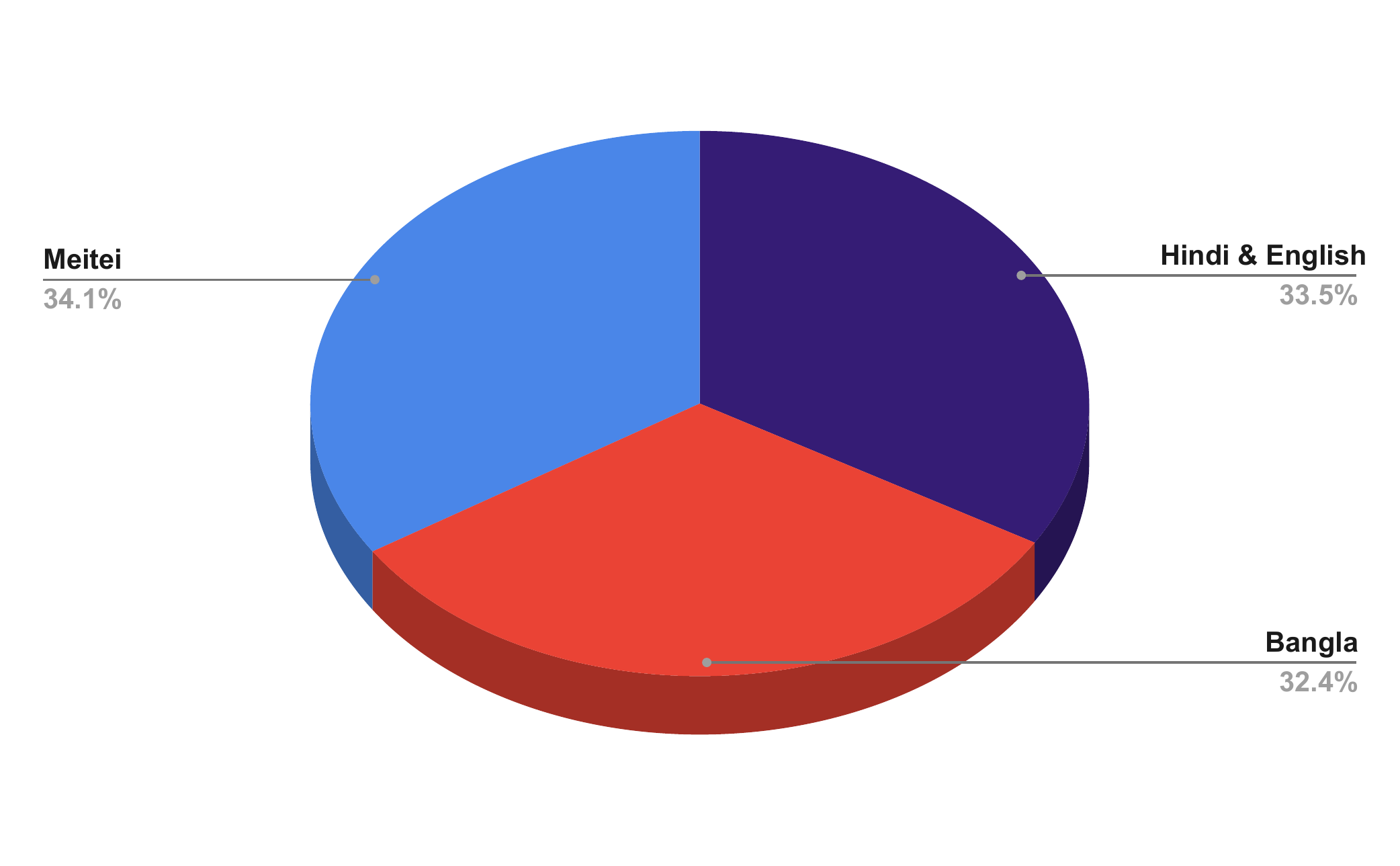}
\caption{Test Dataset Proportion}
\label{test_data}
\end{figure}

Test data consists total of around 3000 comments, precisely 34.1\% Meitei, 32.4\% Bangla, and 33.5\% Hindi shown in Figure \ref{test_data}.

The sources for the test data comments are YouTube, Twitter and Telegram. The distribution of aggressive comments in the test data given in the Table \ref{testdataset}, in Meitei most of the comments are `covertly aggressive'. The `overtly aggressive'  and `non - aggressive' are in equal proportion.
Bangla has `covertly aggressive' and 'non - aggressive' comments in equal proportion and `overtly aggressive' are more in number. Hindi has least number of `covertly aggressive' comments. 

\begin{table}[!h]
\begin{center}
\begin{tabular}{l|c|ccc}
\cline{1-5}
\multirow{2}{*}{} & \multicolumn{4}{c}{\textbf{Aggression}}                   \\ \cline{2-5}
& \textbf{TOTAL} & \multicolumn{1}{l}{\textbf{OAG}} & \textbf{CAG} & \textbf{NAG} \\ \cline{1-5}

\textbf{Meitei}    & \textbf{1,020}   & 315 & 391  & 314  \\ \cline{1-5}

\textbf{Bangla} & \textbf{967}   & 465 & 244 & 258 \\ \cline{1-5}

\textbf{Hindi \& English}  & \textbf{1,002} & 440 & 85 & 477  \\ \cline{1-5}

\textbf{Multilingual}  & \textbf{2,989} & 1,220 & 720 & 1,049\\ \cline{1-5}

\multirow{2}{*}{} & \multicolumn{4}{c}{\textbf{Gendered}} \\
\cline{2-4}

& \textbf{TOTAL} & \multicolumn{1}{l}{\textbf{GEN}} & \textbf{NGEN} \\ \cline{1-4}

\textbf{Meitei}    & \textbf{1,020}   & 317 & 703 \\ \cline{1-4}

\textbf{Bangla}    & \textbf{967}   & 303                            & 664  \\ \cline{1-4}

\textbf{Hindi \& English}  & \textbf{1,002} & 204 & 798 \\ \cline{1-4}

\textbf{Multilingual}  & \textbf{2,989} & 824 & 2,165\\ \cline{1-4}

\multirow{2}{*}{} & \multicolumn{4}{c}{\textbf{Communal}} \\ \cline{2-4}

& \textbf{TOTAL} & \multicolumn{1}{l}{\textbf{COM}} & \textbf{NCOM} \\ \cline{1-4}

\textbf{Meitei}    & \textbf{1,020} & 141 & 879 \\ \cline{1-4}

\textbf{Bangla}    & \textbf{967}   & 106 & 861  \\ \cline{1-4}

\textbf{Hindi \& English}  & \textbf{1,002} & 362 & 640 \\ \cline{1-4}

\textbf{Multilingual}  & \textbf{2,989} & 609 & 2,380\\ \cline{1-4}

\end{tabular}
\caption{The ICON Test Dataset}\label{testdataset}
\end{center}
\end{table}

\begin{figure}[!h]
\centering
\includegraphics[width=\columnwidth]{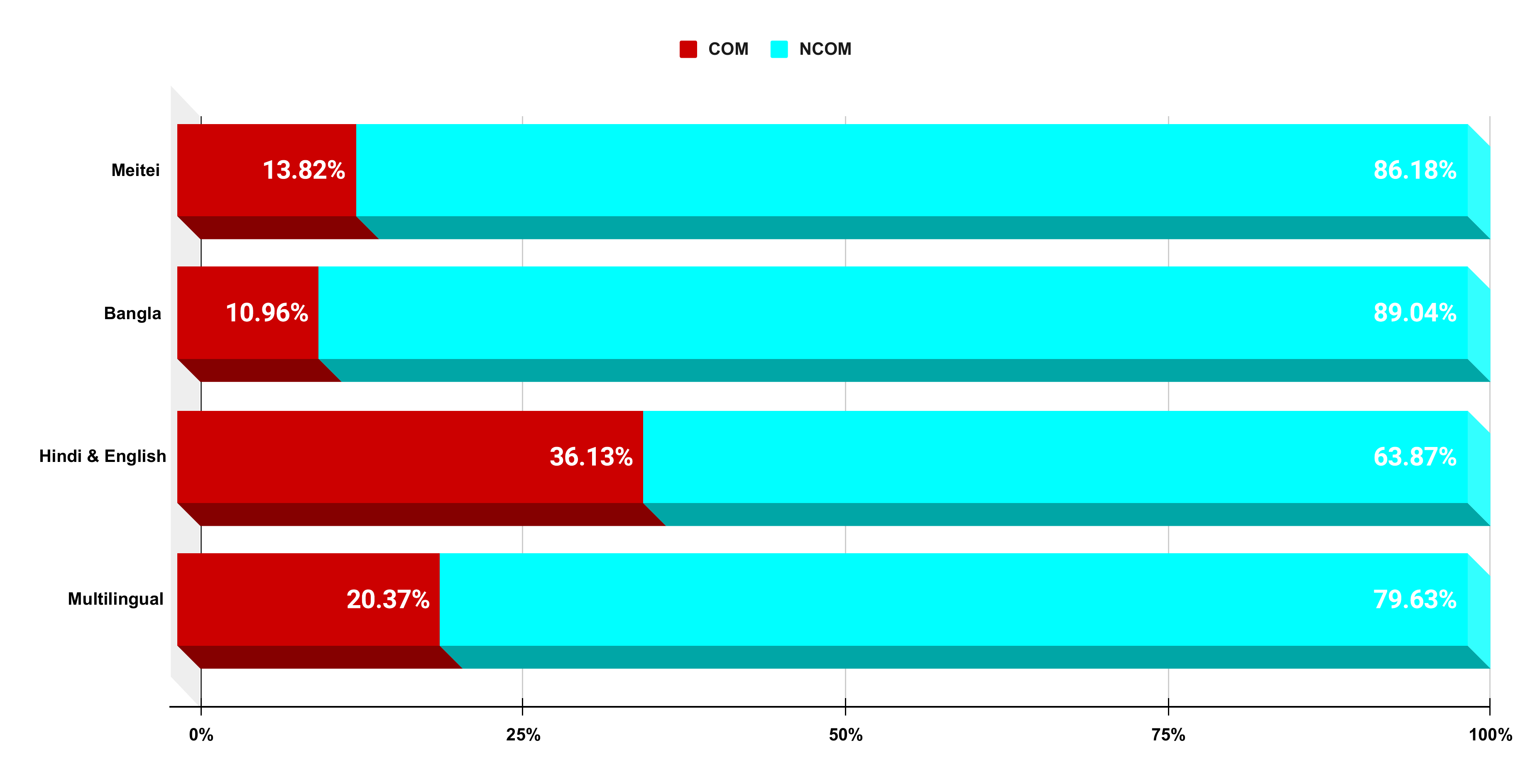}
\caption{Communal Test Dataset}
\label{testcom}
\end{figure}

Overall 28\% of the comments are gendered and 20\% comments are communal. Hindi has most number of communal comments given in Figure \ref{testcom}. 

\begin{figure}[!h]
\centering
\includegraphics[width=\columnwidth]{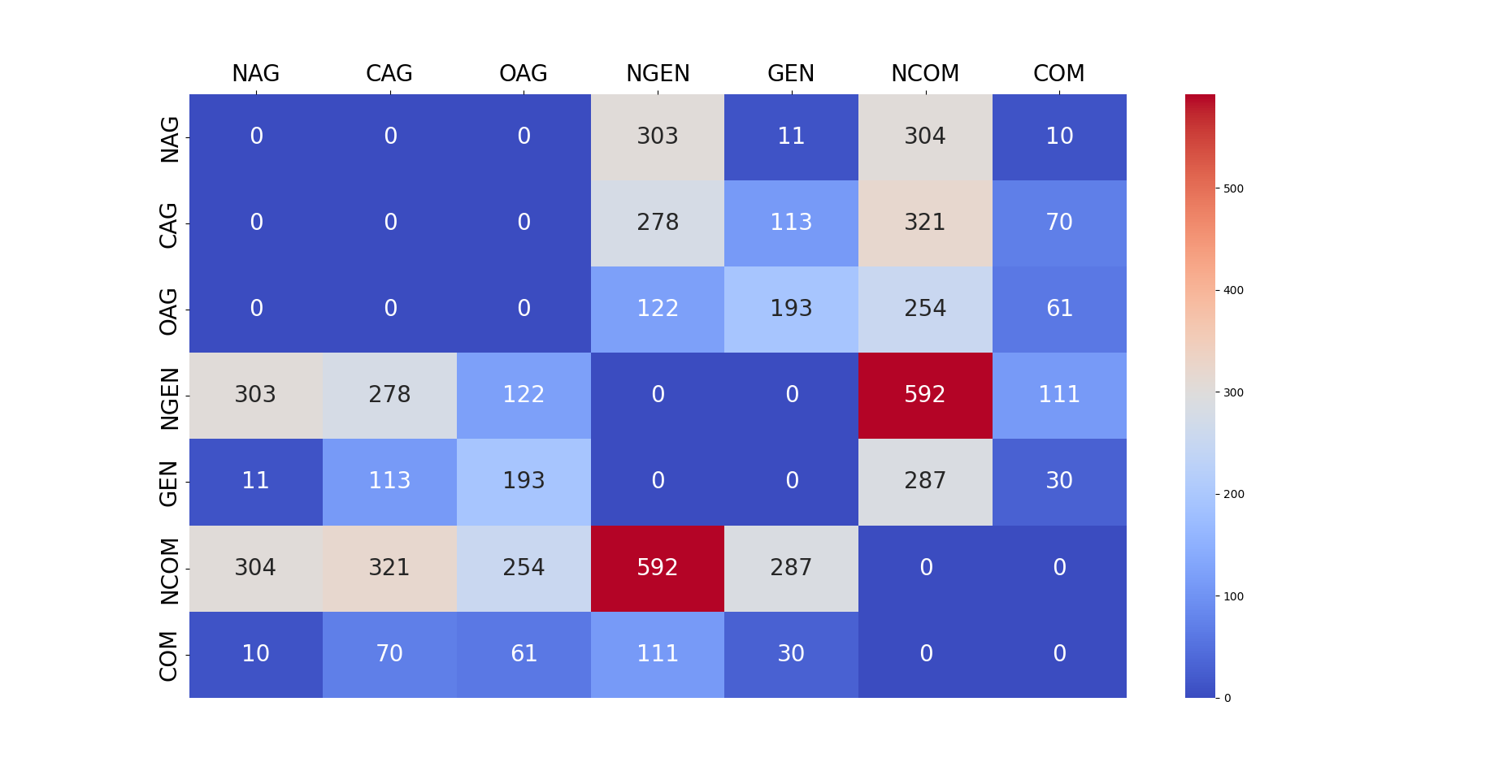}
\caption{Co-occurrence heatmap for Meitei Test Dataset}
\label{testheatmapmni}
\end{figure}

\begin{figure}[!h]
\centering
\includegraphics[width=\columnwidth]{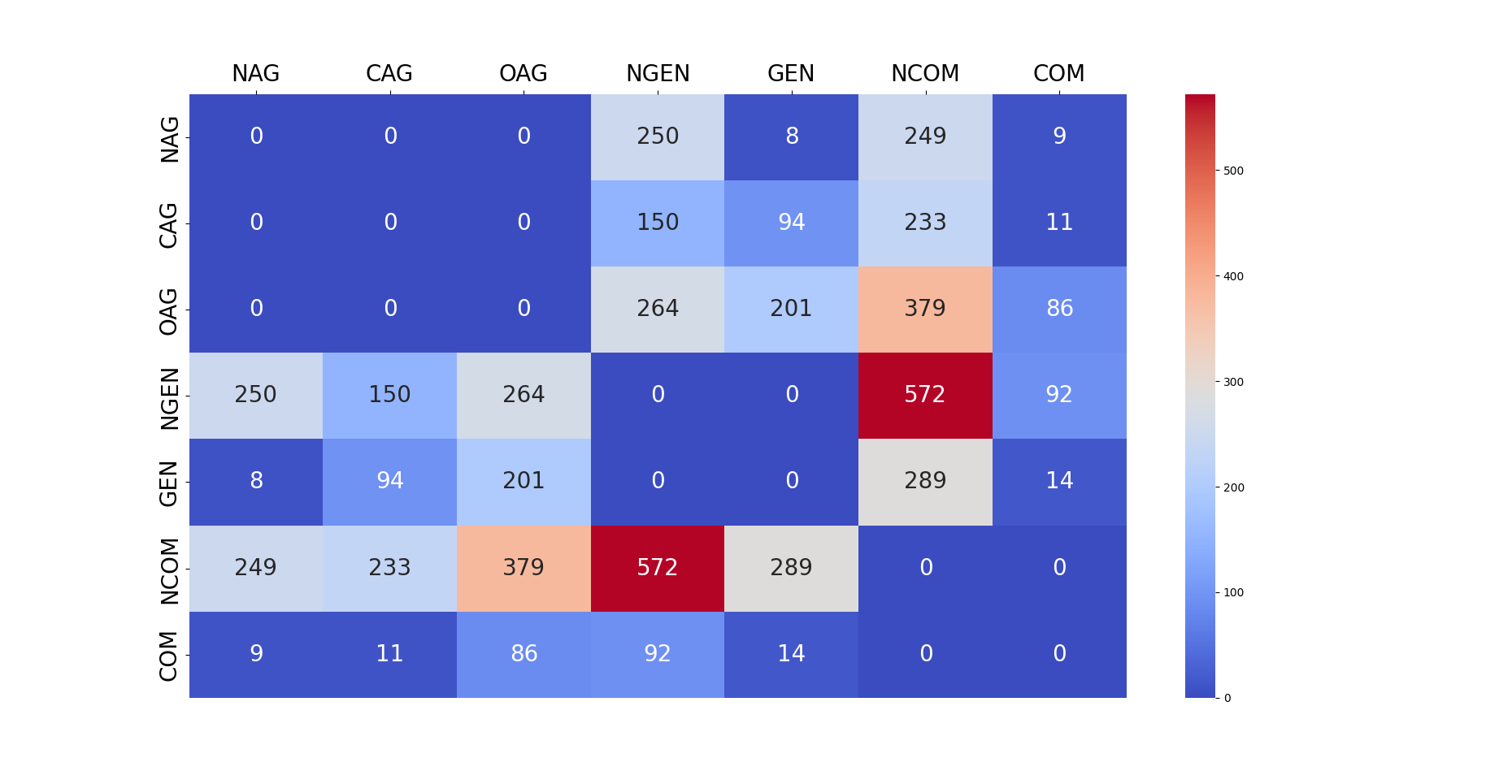}
\caption{Co-occurrence heatmap  for Bangla Test Dataset}
\label{testheatmapben}
\end{figure}

\begin{figure}[!h]
\centering
\includegraphics[width=\columnwidth]{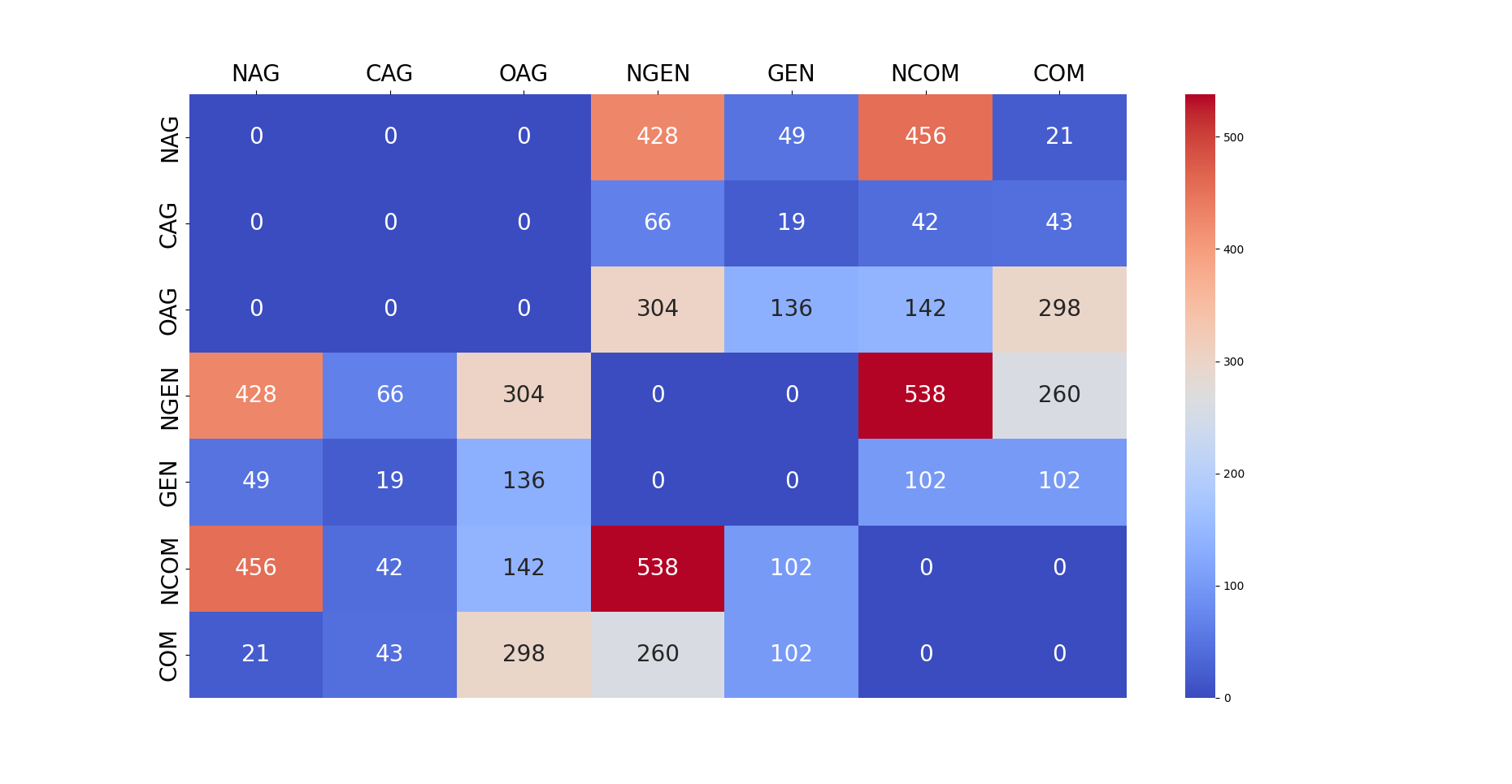}
\caption{Co-occurrence heatmap for Hindi Test Dataset}
\label{testheatmaphin}
\end{figure}

\begin{figure}[!h]
\centering
\includegraphics[width=\columnwidth]{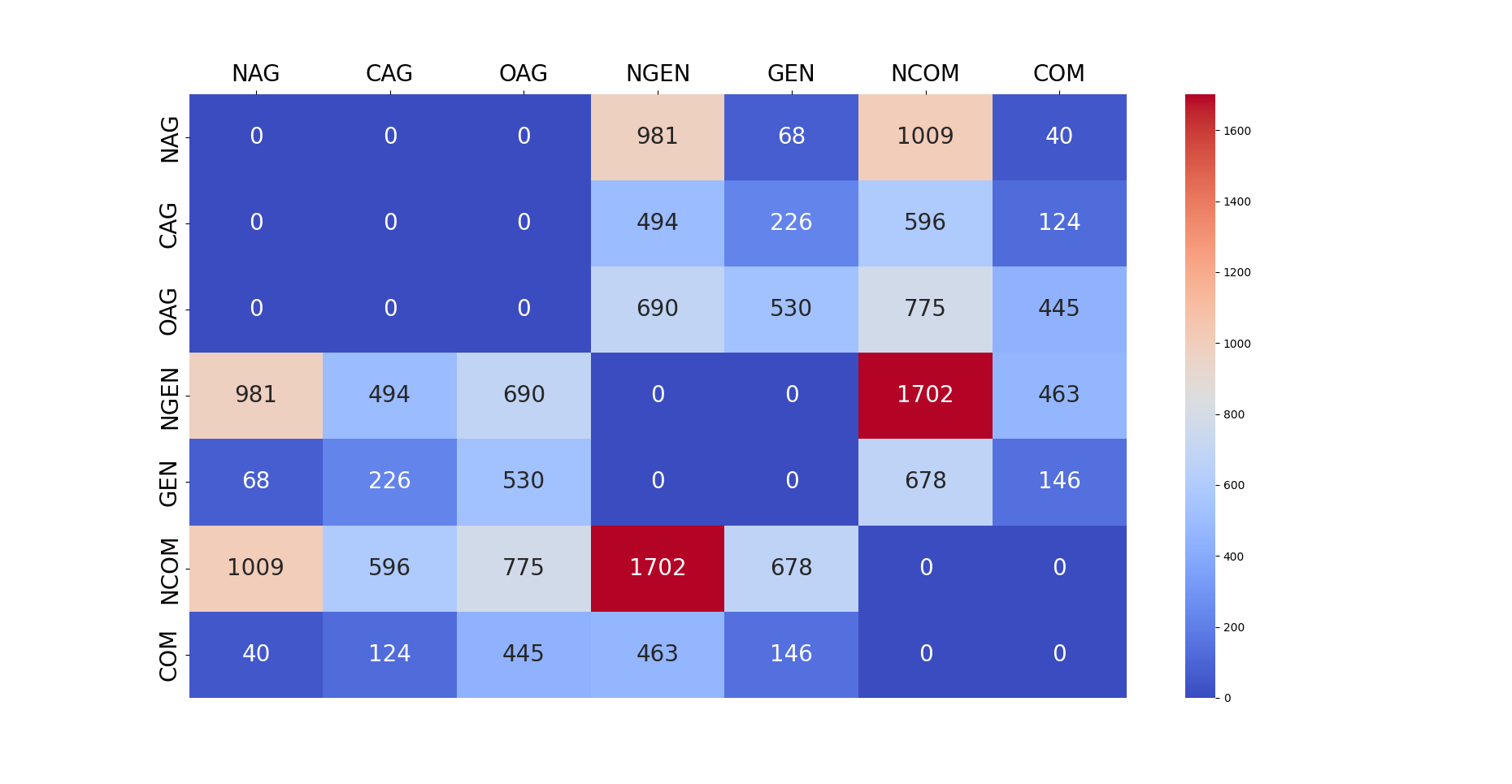}
\caption{Co-occurrence heatmap In Test Dataset}
\label{testheatmapalllang}
\end{figure}

If we look at co-occurence of different tags in the Meitei test data illustrated in the Figure \ref{testheatmapmni}, out of 141 communal comments of Meitei 131 are aggressive, and out of 317 gendered comments 306 has aggression.
In Figure \ref{testheatmapben}, Bangla test set has 97 communal comments which are also aggressive, out of total 106 communal comments and 295 out of 303 gendered comments are aggressive.
In Hindi test data set co-occurence of the tags shown in the Figure \ref{testheatmaphin}, 102 out of 204 gendered comments are communal and 341 out of 362 communal comments have some kind of aggression. 
From Figure \ref{testheatmapalllang}, we can say that most of the gendered and communal comments are `overtly aggressive'.

\begin{figure}[!h]
\centering
\includegraphics[width=\columnwidth]{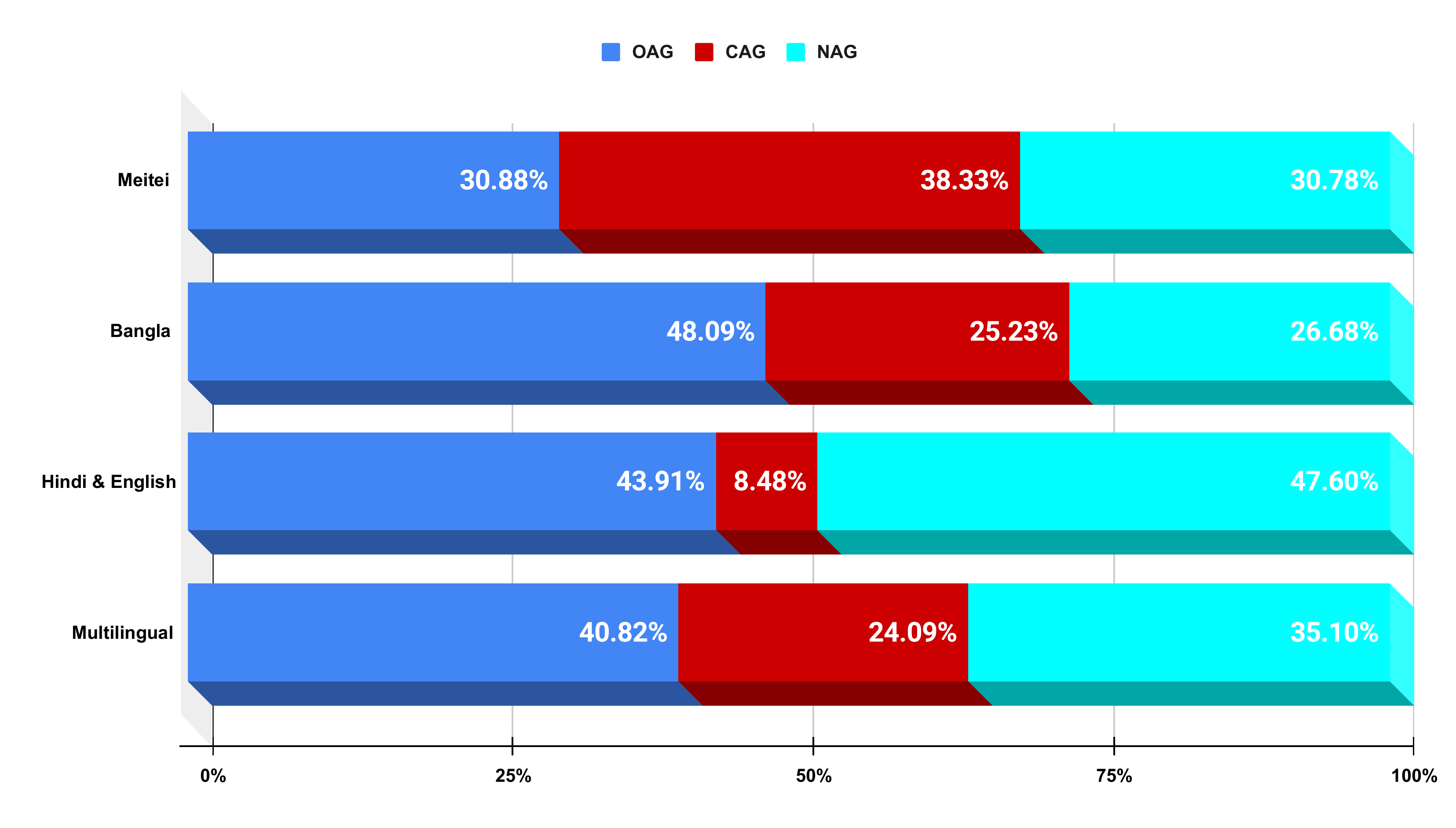}
\caption{Aggression Test Dataset}
\label{testaggr}
\end{figure}

\begin{figure}[!h]
\centering
\includegraphics[width=\columnwidth]{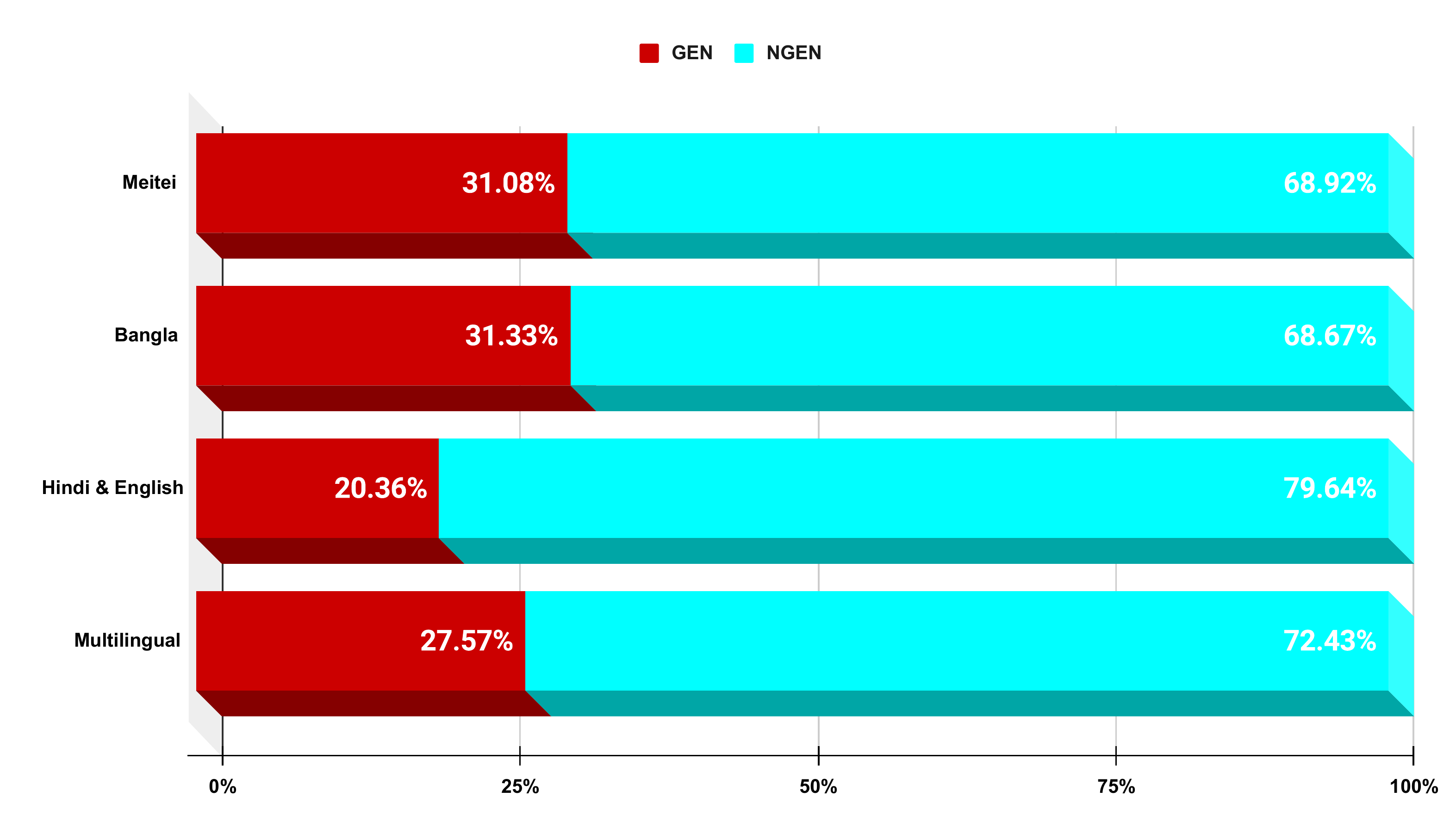}
\caption{Misogyny Test Dataset}
\label{testmiso}
\end{figure}

\section{Baseline Experiments and Results}
We have tried different basic preprocessing of the train data such as lowercasing of characters, remove stopwords of english available in NLTK, hindi and bangla stopwords taken from\footnote{https://github.com/WorldBrain/remove-stopwords/tree/master/lib}, replacing 'url', '@someusername', with 'URL', '@username' respectively and remove '\#' in '\#tags'. Improvement in results could only be seen when character caselowering was done and stopwords have been removed, all other preprocessing and their combination did not improve the results.
We have used combination of character[3, 4, 5] and word[1, 2, 3] n-gram as features and tf-idf as weight scheme to train SVM with different C values [0.001, 0.01, 0.1, 1, 5, 10].
Training and testing of the model is done on train and test data set details given in Table \ref{traindataset}  and Table \ref{testdataset} respectively.
Evaluation of the model is done using two standard evaluation metrics used for multi-label classification.
\textbf{instance-F1}: It is the F-measure averaging on each instance in the test set i.e. the classification will be considered right only when all the labels in a given instance is predicted correctly. 
\textbf{micro-F1}: It gives a weighted average score of each class and is generally considered a good metric in cases of class-imbalance.
Results for all languages and label class (category) is shown in Table \ref{baselineresulttable}.

\begin{table}[!h]
    \begin{center}
    \small\addtolength{\tabcolsep}{-2pt}
    \begin{tabular}{c|c|c|c}
    \hline
    \textbf{Language} & \textbf{Category} & \textbf{micro-F1} & \textbf{Instance-F1}\\
    \hline
    & Aggression & 0.47  & -\\
    \cline{2-3}
    & Gender Bias & 0.70  & -\\
    \cline{2-3}
    Meitei & Communal Bias & 0.86  & -\\
    \cline{2-3}
    & Overall & 0.68 & -\\
    \cline{2-3}
    & - & - & 0.34 \\
    \hline
    & Aggression & 0.40  & -\\
    \cline{2-3}
    & Gender Bias & 0.75  & -\\
    \cline{2-3}
    Bangla & Communal Bias & 0.88  & -\\
    \cline{2-3}
    & Overall & 0.68 & -\\
    \cline{2-3}
    & - & - & 0.26 \\
    \hline
    & Aggression & 0.56  & -\\
    \cline{2-3}
    & Gender Bias & 0.80  & -\\
    \cline{2-3}
    Hindi & Communal Bias & 0.68  & -\\
    \cline{2-3}
    & Overall & 0.68 & -\\
    \cline{2-3}
    & - & - & 0.26 \\
    \hline
    & Aggression & 0.47  & -\\
    \cline{2-3}
    & Gender Bias & 0.77  & -\\
    \cline{2-3}
    Multilingual & Communal Bias & 0.82  & -\\
    \cline{2-3}
    & Overall & 0.68 & -\\
    \cline{2-3}
    & - & - & 0.29 \\
    \hline
    \end{tabular}
    \caption{Baseline Results}
    \label{baselineresulttable}
    \end{center}
\end{table}

\section{The Way Ahead}

In this paper we have presented a multilingual dataset annotated with different levels of annotation and bias. Currently the dataset consists of over 15,000 data points in Meitei, Bangla, Hindi and English, collected from various sources including YouTube comments, Facebook, Twitter and Telegram.

We are currently working on augmenting the dataset with more data points from more Indian languages from different language families and also from more sources. We are also experimenting with different kinds of models for building an automatic identification system for aggression and bias.

    
    
    

\section{Acknowledgments}

This research is funded by Facebook Research under its Content Policy Research Initiative.

We would like to thank Mohit Raj, Shiladitya Bhattacharya, Afrida Aainun Murshida, Sanju Pukhrambam and Diana Thingujam for helping us out with annotations at different points in the project.

\section{Bibliographical References}\label{reference}

\bibliographystyle{lrec2022-bib}
\bibliography{references}

\begin{thebibliography}{}

\bibitem[\protect\citename{Agha}2006]{Agha2006}
Agha, A.
\newblock (2006).
\newblock {\em Language and Social Relations}.
\newblock Cambridge: Cambridge University Press.

\bibitem[\protect\citename{Akhtar \bgroup et al.\egroup }2019]{Akhtar2019}
Akhtar, S., Basile, V., and Patti, V., (2019).
\newblock {\em A New Measure of Polarization in the Annotation of Hate Speech},
  pages 588--603.
\newblock 11.

\bibitem[\protect\citename{Al-Hassan and Al-Dossari}2019]{Al-Hassan2019}
Al-Hassan, A. and Al-Dossari, H.
\newblock (2019).
\newblock Detection of hate speech in social networks: A survey on multilingual
  corpus.
\newblock pages 83--100, 02.

\bibitem[\protect\citename{Albadi \bgroup et al.\egroup }2018]{Albadi2018}
Albadi, N., Kurdi, M., and Mishra, S.
\newblock (2018).
\newblock Are they our brothers? analysis and detection of religious hate
  speech in the arabic twittersphere.
\newblock page 69–76, 08.

\bibitem[\protect\citename{Alfina \bgroup et al.\egroup }2017]{Alfina2017}
Alfina, I., Mulia, R., Fanany, M.~I., and Ekanata, Y.
\newblock (2017).
\newblock Hate speech detection in the indonesian language: A dataset and
  preliminary study.
\newblock 10.

\bibitem[\protect\citename{Bhattacharya \bgroup et al.\egroup
  }2020]{bhattacharya-etal-2020-developing}
Bhattacharya, S., Singh, S., Kumar, R., Bansal, A., Bhagat, A., Dawer, Y.,
  Lahiri, B., and Ojha, A.~K.
\newblock (2020).
\newblock Developing a multilingual annotated corpus of misogyny and
  aggression.
\newblock In {\em Proceedings of the Second Workshop on Trolling, Aggression
  and Cyberbullying}, pages 158--168, Marseille, France, May. European Language
  Resources Association (ELRA).

\bibitem[\protect\citename{Bohra \bgroup et al.\egroup
  }2018]{bohra-etal-2018-dataset}
Bohra, A., Vijay, D., Singh, V., Akhtar, S.~S., and Shrivastava, M.
\newblock (2018).
\newblock A dataset of {H}indi-{E}nglish code-mixed social media text for hate
  speech detection.
\newblock In {\em Proceedings of the Second Workshop on Computational Modeling
  of People{'}s Opinions, Personality, and Emotions in Social Media}, pages
  36--41, New Orleans, Louisiana, USA, June. Association for Computational
  Linguistics.

\bibitem[\protect\citename{Chen \bgroup et al.\egroup }2012]{Chen2012}
Chen, Y., Zhou, Y., Zhu, S., and Xu, H.
\newblock (2012).
\newblock Detecting offensive language in social media to protect adolescent
  online safety.
\newblock pages 71--80, 09.

\bibitem[\protect\citename{Chung \bgroup et al.\egroup
  }2019]{chung-etal-2019-conan}
Chung, Y.-L., Kuzmenko, E., Tekiroglu, S.~S., and Guerini, M.
\newblock (2019).
\newblock {CONAN} - {CO}unter {NA}rratives through nichesourcing: a
  multilingual dataset of responses to fight online hate speech.
\newblock In {\em Proceedings of the 57th Annual Meeting of the Association for
  Computational Linguistics}, pages 2819--2829, Florence, Italy, July.
  Association for Computational Linguistics.

\bibitem[\protect\citename{Davidson \bgroup et al.\egroup }2017]{Davidson2017}
Davidson, T., Warmsley, D., Macy, M., and Weber, I.
\newblock (2017).
\newblock Automated hate speech detection and the problem of offensive
  language.
\newblock 03.

\bibitem[\protect\citename{de Pelle and Moreira}2016]{dePelle2016}
de~Pelle, R. and Moreira, V.~P.
\newblock (2016).
\newblock Offensive comments in the brazilian web: A dataset and baseline
  results.
\newblock page 510–519.

\bibitem[\protect\citename{Fortuna \bgroup et al.\egroup
  }2019]{fortuna-etal-2019-hierarchically}
Fortuna, P., Rocha~da Silva, J., Soler-Company, J., Wanner, L., and Nunes, S.
\newblock (2019).
\newblock A hierarchically-labeled {P}ortuguese hate speech dataset.
\newblock In {\em Proceedings of the Third Workshop on Abusive Language
  Online}, pages 94--104, Florence, Italy, August. Association for
  Computational Linguistics.

\bibitem[\protect\citename{Haddad \bgroup et al.\egroup }2019]{Haddad2019}
Haddad, H., Mulki, H., and Oueslati, A., (2019).
\newblock {\em T-HSAB: A Tunisian Hate Speech and Abusive Dataset}, pages
  251--263.
\newblock 10.

\bibitem[\protect\citename{Hammer}2017]{Hammer2017}
Hammer, H.
\newblock (2017).
\newblock Automatic detection of hateful comments in online discussion.
\newblock volume 188, pages 164--173, 01.

\bibitem[\protect\citename{Jurgens \bgroup et al.\egroup
  }2019]{jurgens-etal-2019-just}
Jurgens, D., Hemphill, L., and Chandrasekharan, E.
\newblock (2019).
\newblock A just and comprehensive strategy for using {NLP} to address online
  abuse.
\newblock In {\em Proceedings of the 57th Annual Meeting of the Association for
  Computational Linguistics}, pages 3658--3666, Florence, Italy, July.
  Association for Computational Linguistics.

\bibitem[\protect\citename{Kaggle}2020]{Kaggle2020}
Kaggle.
\newblock (2020).
\newblock Jigsaw multilingual toxic comment classification.

\bibitem[\protect\citename{Kolhatkar \bgroup et al.\egroup
  }2020]{Kolhatkar2020}
Kolhatkar, V., Wu, H., Cavasso, L., Francis, E., Shukla, K., and Taboada, M.
\newblock (2020).
\newblock The sfu opinion and comments corpus: A corpus for the analysis of
  online news comments.
\newblock {\em Corpus Pragmatics}, 4, 06.

\bibitem[\protect\citename{Kumar \bgroup et al.\egroup
  }2018a]{kumar-etal-2018-benchmarking}
Kumar, R., Ojha, A.~K., Malmasi, S., and Zampieri, M.
\newblock (2018a).
\newblock Benchmarking aggression identification in social media.
\newblock In {\em Proceedings of the First Workshop on Trolling, Aggression and
  Cyberbullying ({TRAC}-2018)}, pages 1--11, Santa Fe, New Mexico, USA, August.
  Association for Computational Linguistics.

\bibitem[\protect\citename{Kumar \bgroup et al.\egroup
  }2018b]{kumar-etal-2018-aggression}
Kumar, R., Reganti, A.~N., Bhatia, A., and Maheshwari, T.
\newblock (2018b).
\newblock Aggression-annotated corpus of {H}indi-{E}nglish code-mixed data.
\newblock In {\em Proceedings of the Eleventh International Conference on
  Language Resources and Evaluation ({LREC} 2018)}, Miyazaki, Japan, May.
  European Language Resources Association (ELRA).

\bibitem[\protect\citename{Kumar \bgroup et al.\egroup
  }2020]{kumar-etal-2020-evaluating}
Kumar, R., Ojha, A.~K., Malmasi, S., and Zampieri, M.
\newblock (2020).
\newblock Evaluating aggression identification in social media.
\newblock In {\em Proceedings of the Second Workshop on Trolling, Aggression
  and Cyberbullying}, pages 1--5, Marseille, France, May. European Language
  Resources Association (ELRA).

\bibitem[\protect\citename{Malmasi and
  Zampieri}2017]{malmasi-zampieri-2017-detecting}
Malmasi, S. and Zampieri, M.
\newblock (2017).
\newblock Detecting hate speech in social media.
\newblock In {\em Proceedings of the International Conference Recent Advances
  in Natural Language Processing, {RANLP} 2017}, pages 467--472, Varna,
  Bulgaria, September. INCOMA Ltd.

\bibitem[\protect\citename{Mandl \bgroup et al.\egroup }2020]{Mandl2020}
Mandl, T., Modha, S., Shahi, G.~K., Jaiswal, A., Nandini, D., Patel, D.,
  Majumder, P., and Schäfer, J.
\newblock (2020).
\newblock Overview of the hasoc track at fire 2020: Hate speech and offensive
  content identification in indo-european languages.
\newblock page 29–32, 12.

\bibitem[\protect\citename{Martins \bgroup et al.\egroup }2018]{Martins2018}
Martins, R., Gomes, M., Almeida, J., Novais, P., and Henriques, P.
\newblock (2018).
\newblock Hate speech classification in social media using emotional analysis.
\newblock pages 61--66, 10.

\bibitem[\protect\citename{Mathur \bgroup et al.\egroup
  }2018]{mathur-etal-2018-detecting}
Mathur, P., Shah, R., Sawhney, R., and Mahata, D.
\newblock (2018).
\newblock Detecting offensive tweets in {H}indi-{E}nglish code-switched
  language.
\newblock In {\em Proceedings of the Sixth International Workshop on Natural
  Language Processing for Social Media}, pages 18--26, Melbourne, Australia,
  July. Association for Computational Linguistics.

\bibitem[\protect\citename{Mubarak \bgroup et al.\egroup
  }2017]{mubarak-etal-2017-abusive}
Mubarak, H., Darwish, K., and Magdy, W.
\newblock (2017).
\newblock Abusive language detection on {A}rabic social media.
\newblock In {\em Proceedings of the First Workshop on Abusive Language
  Online}, pages 52--56, Vancouver, BC, Canada, August. Association for
  Computational Linguistics.

\bibitem[\protect\citename{Nascimento \bgroup et al.\egroup
  }2019]{Nascimento2019}
Nascimento, G., Carvalho, F., Cunha, A., Viana, C., and Paiva~Guedes, G.
\newblock (2019).
\newblock Hate speech detection using brazilian imageboards.
\newblock pages 325--328, 10.

\bibitem[\protect\citename{Nobata \bgroup et al.\egroup }2016]{Nobata2016}
Nobata, C., Tetreault, J., Thomas, A., Mehdad, Y., and Chang, Y.
\newblock (2016).
\newblock Abusive language detection in online user content.
\newblock pages 145--153, 04.

\bibitem[\protect\citename{Ousidhoum \bgroup et al.\egroup
  }2019]{ousidhoum-etal-2019-multilingual}
Ousidhoum, N., Lin, Z., Zhang, H., Song, Y., and Yeung, D.-Y.
\newblock (2019).
\newblock Multilingual and multi-aspect hate speech analysis.
\newblock In {\em Proceedings of the 2019 Conference on Empirical Methods in
  Natural Language Processing and the 9th International Joint Conference on
  Natural Language Processing (EMNLP-IJCNLP)}, pages 4675--4684, Hong Kong,
  China, November. Association for Computational Linguistics.

\bibitem[\protect\citename{Poletto \bgroup et al.\egroup
  }2021]{Poletto2021ResourcesAB}
Poletto, F., Basile, V., Sanguinetti, M., Bosco, C., and Patti, V.
\newblock (2021).
\newblock Resources and benchmark corpora for hate speech detection: a
  systematic review.
\newblock {\em Lang. Resour. Evaluation}, 55:477--523.

\bibitem[\protect\citename{Ross \bgroup et al.\egroup }2017]{Ross2017}
Ross, B., Rist, M., Carbonell, G., Cabrera, B., Kurowsky, N., and Wojatzki, M.
\newblock (2017).
\newblock Measuring the reliability of hate speech annotations: The case of the
  european refugee crisis.
\newblock In {\em NLP4CMC III: 3rd workshop on natural language processing for
  computer-mediated communication.}, 09.

\bibitem[\protect\citename{Sch{\"a}fer and
  Burtenshaw}2019]{schafer-burtenshaw-2019-offence}
Sch{\"a}fer, J. and Burtenshaw, B.
\newblock (2019).
\newblock Offence in dialogues: A corpus-based study.
\newblock In {\em Proceedings of the International Conference on Recent
  Advances in Natural Language Processing (RANLP 2019)}, pages 1085--1093,
  Varna, Bulgaria, September. INCOMA Ltd.

\bibitem[\protect\citename{Schmidt and
  Wiegand}2017]{schmidt-wiegand-2017-survey}
Schmidt, A. and Wiegand, M.
\newblock (2017).
\newblock A survey on hate speech detection using natural language processing.
\newblock In {\em Proceedings of the Fifth International Workshop on Natural
  Language Processing for Social Media}, pages 1--10, Valencia, Spain, April.
  Association for Computational Linguistics.

\bibitem[\protect\citename{Vidgen and Derczynski}2020]{Vidgen2020}
Vidgen, B. and Derczynski, L.
\newblock (2020).
\newblock Directions in abusive language training data, a systematic review:
  Garbage in, garbage out.
\newblock {\em PLOS ONE}, 15:e0243300, 12.

\bibitem[\protect\citename{Vidgen and Yasseri}2020]{Vidgen2020Detectingweak}
Vidgen, B. and Yasseri, T.
\newblock (2020).
\newblock Detecting weak and strong islamophobic hate speech on social media.
\newblock {\em Journal of Information Technology \& Politics}, 17:66--78, 01.

\bibitem[\protect\citename{Waseem and Hovy}2016]{waseem-hovy-2016-hateful}
Waseem, Z. and Hovy, D.
\newblock (2016).
\newblock Hateful symbols or hateful people? predictive features for hate
  speech detection on {T}witter.
\newblock In {\em Proceedings of the {NAACL} Student Research Workshop}, pages
  88--93, San Diego, California, June. Association for Computational
  Linguistics.

\bibitem[\protect\citename{Waseem \bgroup et al.\egroup
  }2017]{waseem-etal-2017-understanding}
Waseem, Z., Davidson, T., Warmsley, D., and Weber, I.
\newblock (2017).
\newblock Understanding abuse: A typology of abusive language detection
  subtasks.
\newblock In {\em Proceedings of the First Workshop on Abusive Language
  Online}, pages 78--84, Vancouver, BC, Canada, August. Association for
  Computational Linguistics.

\bibitem[\protect\citename{Waseem}2016]{Waseem2016}
Waseem, Z.
\newblock (2016).
\newblock Are you a racist or am i seeing things? annotator influence on hate
  speech detection on twitter.
\newblock pages 138--142, 01.

\bibitem[\protect\citename{Weingartner and
  Stahel}2019]{weingartner-stahel-2019-online}
Weingartner, S. and Stahel, L.
\newblock (2019).
\newblock Online aggression from a sociological perspective: An integrative
  view on determinants and possible countermeasures.
\newblock In {\em Proceedings of the Third Workshop on Abusive Language
  Online}, pages 181--187, Florence, Italy, August. Association for
  Computational Linguistics.

\bibitem[\protect\citename{Zampieri \bgroup et al.\egroup
  }2019a]{zampieri-etal-2019-predicting}
Zampieri, M., Malmasi, S., Nakov, P., Rosenthal, S., Farra, N., and Kumar, R.
\newblock (2019a).
\newblock Predicting the type and target of offensive posts in social media.
\newblock In {\em Proceedings of the 2019 Conference of the North {A}merican
  Chapter of the Association for Computational Linguistics: Human Language
  Technologies, Volume 1 (Long and Short Papers)}, pages 1415--1420,
  Minneapolis, Minnesota, June. Association for Computational Linguistics.

\bibitem[\protect\citename{Zampieri \bgroup et al.\egroup
  }2019b]{zampieri-etal-2019-semeval}
Zampieri, M., Malmasi, S., Nakov, P., Rosenthal, S., Farra, N., and Kumar, R.
\newblock (2019b).
\newblock {S}em{E}val-2019 task 6: Identifying and categorizing offensive
  language in social media ({O}ffens{E}val).
\newblock In {\em Proceedings of the 13th International Workshop on Semantic
  Evaluation}, pages 75--86, Minneapolis, Minnesota, USA, June. Association for
  Computational Linguistics.

\bibitem[\protect\citename{Zampieri \bgroup et al.\egroup
  }2020]{zampieri-etal-2020-semeval}
Zampieri, M., Nakov, P., Rosenthal, S., Atanasova, P., Karadzhov, G., Mubarak,
  H., Derczynski, L., Pitenis, Z., and {\c{C}}{\"o}ltekin, {\c{C}}.
\newblock (2020).
\newblock {S}em{E}val-2020 task 12: Multilingual offensive language
  identification in social media ({O}ffens{E}val 2020).
\newblock In {\em Proceedings of the Fourteenth Workshop on Semantic
  Evaluation}, pages 1425--1447, Barcelona (online), December. International
  Committee for Computational Linguistics.

\end{thebibliography}


\bibliographystylelanguageresource{lrec2022-bib}

\appendix
\section{Data Statement}

\subsection{Header}


\begin{itemize}
\item[] \textsl{Dataset Title: ComMA Dataset v0.2}
\item[] \textsl{Dataset Curator(s): }

\begin{itemize}
    \item \textbf{Akash Bhagat}, Indian Institute of Technology-Kharagpur
    
    \item \textbf{Enakshi Nandi}, Panlingua Language Processing LLP, New Delhi
    
    \item \textbf{Laishram Niranjana Devi}, Panlingua Language Processing LLP, New Delhi
    
    \item \textbf{Mohit Raj}, Panlingua Language Processing LLP, New Delhi
    
    \item \textbf{Shyam Ratan}, Dr. Bhimrao Ambedkar University, Agra
    
    \item \textbf{Siddharth Singh}, Dr. Bhimrao Ambedkar University, Agra
    
    \item \textbf{Yogesh Dawer}, Dr. Bhimrao Ambedkar University, Agra
    
\end{itemize}

\item[] \textsl{Dataset Version: Version 0.2, 2nd October 2021}

\item[] \textsl{Dataset Citation: NA}

\item[] \textsl{Data Statement Authors: }

\begin{itemize}
\item \textbf{Enakshi Nandi}, Panlingua Language Processing LLP, New Delhi
    
\item \textbf{Laishram Niranjana Devi}, Panlingua Language Processing LLP, New Delhi
    
\item \textbf{Shyam Ratan}, Dr. Bhimrao Ambedkar University
\end{itemize}

\item[] \textsl{Data Statement Version: 1, 17th November 2021}

\item[] \textsl{Data Statement Citation and DOI:} NA

\item[] \textsl{Links to versions of this data statement in other languages: NA}

\end{itemize}

\subsection{Executive Summary}

The objective of working on this dataset is to identify and tag aggression and various kinds of bias (gender, communal, caste/class, ethnic/racial) in social media discourse. To that end, this dataset has been compiled by collecting over 15,000 comments from YouTube, Facebook, Twitter and Telegram in Meitei, Bangla, Hindi and English, with around 5000 comments in Meitei, Bangla and Hindi each, which were all mixed with English data. The data was collected from videos and posts that were politically, socially, sexually, religiously, racially, or otherwise polarized or controversial in nature, so as to elicit a wide and extensive range of hateful, aggressive, gendered, communal, casteist, classist, and racist speech data for our dataset. 


\subsection{Curation Rationale}


This dataset was created with the ultimate goal of developing a system that is able to identify and tag aggression, gender bias, communal bias, caste/class bias, and ethnic/racial bias in social media discourse. To that end, this dataset has been manually annotated by multiple annotators in order to identify the linguistic and pragmatic features that characterize aggression, gender bias, communal bias, caste/class bias, and ethnic/racial bias in the comments on posts, videos, and articles posted on social media sites such as YouTube, Facebook, Twitter, and Telegram.

The specific social media posts and articles whose comments we collected were selected manually, and then crawled with the help of their respective web crawlers. This selection process was contingent on many factors, the chief of which was the need to collect as many aggressive, gender biased, communal, casteist, classist, and racist comments as possible to create a robust dataset. To that end, we focused on identifying controversial posts of a politically, socially, sexually, communally, and racially charged nature that have elicited a significant number of the kind of comments described above. We have then followed similar suggested posts, videos, or articles on the platform to collect more data of a similar or comparable nature. The second factor was language: the comments needed to be in Meitei, Bangla, and Hindi for the most part, with English comments included because they are ubiquitous in the context of Indian social media. 

The dataset was organized in the form of a spreadsheet, with each comment identified by a unique comment code that would help the annotators distinguish an independent comment from comments featuring in a thread, and each comment posted under an article or video constituting one data instance, regardless of its length, language, or content. In other words, a data instance can be a single letter or an essay-length comment, can be written in a single language or a combination of languages, and can contain text (in any script), numerals, and emojis individually or all in one comment. The data instance is annotated taking the entire comment as one single, compact unit.   

\subsection{Documentation for Source Datasets}
This dataset has been developed from a source dataset that marked only aggressive speech collected from public Facebook and Twitter pages, and a subsequent source dataset that developed the tagset to include speech with aggression and gender bias, collected from Facebook, Twitter, and YouTube. 

The links for the research papers published and the workshops conducted on the respective source datasets are listed below:

\begin{enumerate}

    \item \url{http://arxiv.org/abs/1803.09402}
    
    \item Trac - 1\footnote{First Workshop on Trolling, Aggression and Cyberbullying}, 
    \url{https://aclanthology.org/W18-4401} 
    
    \item \url{https://arxiv.org/abs/2003.07428}
    
    \item Trac - 2\footnote{Second Workshop on Trolling, Aggression and Cyberbullying},
    \url{https://aclanthology.org/2020.trac-1.1}
    
\end{enumerate}

The current dataset was built on the foundation laid down by these source datasets, and has added several new, finely-grained tags, including two primary tags marking caste/class bias and ethnic/racial bias, and two secondary tags that mark the discursive roles and discursive effects of (overtly and covertly) aggressive speech.


\subsection{Language Varieties}


The languages included in this dataset, listed with their respective BCP-47 language tags, include:
\begin{itemize}
\item \textbf{code unavailable: Meitei} as spoken by the Meitei community in Manipur, India.
\item \textbf{bn-IN and bn-BD: Bangla} (and its varieties) as spoken in India and Bangladesh. 
\item \textbf{he-IN: Hindi} (and its varieties) as spoken in various parts of India.
\item \textbf{en-IN: English} (and its varieties) as spoken in India, otherwise known as Indian English.
\end{itemize}

Since this dataset has been exclusively collected from online sources, the users writing the comments are assumed to be multilingual and may be based in any part of the world, not just in the places these languages are primarily spoken in. However, the language varieties used in the dataset are primarily those mentioned in the list above.  

\subsection{Speaker Demographic}


This dataset has been sourced exclusively from the internet, hence the speaker demographic of the dataset cannot be identified beyond the language they speak. It is assumed that the speakers could be of any age, gender, sexual orientation, educational background, nationality, caste, class, religion, race, tribe, or ethnicity. 

The speakers are probably multilingual as well, with the language they post in being one of the many they would know or be fluent in. It is a safe assumption to make that many of these comments are made by Indians (specifically people who have Meitei, Bangla, and Hindi as their first or primary language) and Bangladeshis given the nature and reach of the topics selected, but this assumption is not backed by any data or statistical findings.


\subsection{Annotator Demographic}

The annotation scheme and guidelines for this dataset has been developed by Dr Ritesh Kumar, the principal investigator of the ComMA Project and a faculty at the Department of Transdisciplinary Studies, Dr. Bhimrao Ambedkar University, Agra, India. 
He was assisted by the co-PIs of the project - Dr. Bornini Lahiri, Assistant Professor at IIT-Kharagpur and Dr. Atul Kr. Ojha and Akanksha Bansal, co-founders of Panlingua Language Processing LLP - and annotators of this dataset, who have been listed below. Further, these annotators have manually identified the appropriate posts and videos to work on, crawled the data, and then annotated and analysed the processed data in their respective languages.

\begin{itemize}
    
    \item A 31-year-old Bengali Muslim woman working from Gangtok, Sikkim. She has a PhD in English, speaks Bangla, Hindi, and English, and her ideological leanings are centrist. She is annotating the Bangla data.
    
    \item A 29-year-old Bengali Hindu man working from Malda, West Bengal. He has an MA in Linguistics, and speaks Bangla, English, Hindi and Bhojpuri. He is annotating the Bangla data.
    
    \item A 33-year-old Bengali Hindu woman working from Kalyani, West Bengal. She has a PhD in Linguistics, speaks English, Hindi, Bangla, and Sylheti, and her ideological leanings are leftist. She is annotating the Bangla data.
    
    \item A 30-year-old Meitei Hindu woman working from Imphal, Manipur. She is pursuing a PhD in Linguistics, speaks English, Hindi, and Meitei, and her ideological leanings are centrist. She is annotating the Meitei data.
    \item A 28-year-old North Indian Hindu man working from Patna, Bihar. He is pursuing a PhD in Linguistics, speaks English, Hindi, Magahi, and Bhojpuri, and his ideological leanings are centrist. He is annotating the English and Hindi data. 
    
    \item A 25-year-old North Indian Hindu man working from Agra, Uttar Pradesh. He is pursuing an M.Phil in Linguistics, and speaks Braj, Hindi and English. He is annotating the English and Hindi data. 
    
    \item A 27-year-old North Indian Hindu man working from Agra, Uttar Pradesh. He is pursuing an M.Sc. in Computational Linguistics, speaks Hindi, Bhojpuri,and English and his ideological leanings are centrist. He is annotating the English and Hindi data.
    
    \item A 32-year-old Punjabi Hindu man working from Agra, Uttar Pradesh. He is an MA in Journalism and in Linguistics, speaks English, Hindi, and Punjabi, and his ideological leanings are leftist. He is annotating the English and Hindi data. 
    
\end{itemize}


\subsection{Speech Situation and Text Characteristics}

This dataset comprises of online comments written by users of various social media platforms. The comments collected range from 2012 to 2021 (and continuing), and form part of an extensive and intensive social media discourse. 


\begin{itemize}
\item \textbf{Time and place of linguistic activity} - Online
\item \textbf{Date(s) of data collection} - April to September 2021
\item \textbf{Modality} - Written
\item \textbf{Scripted/edited vs. spontaneous} - Spontaneous
\item \textbf{Synchronous vs. asynchronous interaction} - Asynchronous (online comments)    
\item \textbf{Speakers’ intended audience} - Other users of the respective social media platforms and channels 
\item \textbf{Genre} - Social media
\item \textbf{Topic} - Socially or politically polarizing or controversial topics
\item \textbf{Non-linguistic context} - The videos which provide the context for the comments generated
\item \textbf{Additional details about the cultural context} - The sociopolitical climate and cultural context in which the commenters live have a huge influence on the nature, tone, and ideological underpinnnings of the comments they write on social media 
\end{itemize}

\subsection{Preprocessing and Data Formatting}


The preprocessing of the raw data involves deleting all duplicates of a data instance, deleting data instances with urls and texts with less than three words, and removing data instances which occur in languages apart from Meitei, Bangla, Hindi, and English. In the Telegram data, all translations of texts have been deleted manually. The data instances are listed without the names of the commenters, but when someone has replied to a previous comment by tagging them with the '@' symbol, that information is available to the annotator within the text itself.  

Next, the processed data is arranged on a Google spreadsheet, columns are made with the relevant headings and tags (using the option for data validation), and copies of the spreadsheet are shared amongst the annotators working on a particular language so they can annotate the files individually and without consultation with each other. This is to ensure that no annotator is influenced in their annotation by the ideas of another. However, at no stage in the process are the annotators anonymous to each other or anyone else in the team.

\subsection{Capture Quality}

As with any other dataset, we have faced quality issues in data capture. The primary of these is the difficulty in finding every kind of data in every language. For instance, in Bangla, it is very difficult to find racist or communal data, because most conversations that we have come across in social media platforms that are of a communal or racist nature and involves Bangla speakers occur in English. Similar challenges have been faced in Meitei with regard to casteist data, and in Hindi with regard to ethnically and racially biased data. These discrepancies can be explained when the social, political, and cultural contexts of each of these language and speech communities is taken into consideration, which are significantly different from each other.


\subsection{Limitations}

Following the point in the previous section, another limitation in the data is the dearth of comments that can be tagged by the discursive effects of counterspeech, abet and instigate, and gaslighting. In contrast, the discursive effect of attack is very well-represented, not so closely followed by defend. 
All of these factors combine to make it challenging for the dataset in each language to be equally representative of each of the primary tags, thus making it difficult for the researchers to embark on intensive comparative analyses of the characteristics of each of these phenomena across all of the languages being analysed.  

This tagset also does not allow us the option to distinguish a personal attack from an identity based one, to mark national/regional or political bias, and to distinguish sexual harassment from aggression, sexual threat, and gender bias. These are shortcomings that will have to be addressed and resolved in subsequent versions of the tagset.


\subsection{Metadata}





The relevant links to the metadata for this dataset have been provided below:

\begin{itemize}
\item[] \textsl{\textbf{License}: CC BY-NC-SA 4.0} 

\item[] \textsl{\textbf{Annotation Guidelines}:
    \begin{enumerate}
    \item \url{http://arxiv.org/abs/1803.09402} 
    \item \url{https://arxiv.org/abs/2003.07428}
    \item \url{https://drive.google.com/file/d/1ZUZxDaYIfotVur-cJF30cfqIY1dSzJ6K/view?usp=sharing}
    \end{enumerate}}

\item[] \textsl{\textbf{Annotation Process}: Manual annotation} 
\item[] \textsl{\textbf{Dataset Quality Metrics}: Krippendorff's Alpha for IAA} 

\item[] \textsl{\textbf{Errata}: NA}
\end{itemize}

\subsection{Disclosures and Ethical Review}

This dataset has been funded by Facebook Research under Content Policy Research Initiative Phase 2. 

\subsection{Other}

NA

\subsection{Glossary}

NA

\end{document}